\documentclass[11pt,a4paper]{article}
\usepackage[T1]{fontenc}
\usepackage[utf8]{inputenc}
\usepackage{lmodern}
\usepackage[margin=2.5cm]{geometry}
\usepackage{amsmath,amssymb}
\usepackage{graphicx,booktabs,longtable,array}
\usepackage{caption}
\usepackage{setspace}
\usepackage[numbers,sort&compress]{natbib}
\usepackage{microtype}
\usepackage[hidelinks]{hyperref}
\newif\ifanon
\anonfalse 
\ifanon
  \hypersetup{pdftitle={Non-intrusive appliance state detection in unseen homes using classical machine learning},pdfauthor={}}
\else
  \hypersetup{pdftitle={Non-intrusive appliance state detection in unseen homes using classical machine learning},pdfauthor={Zablon Ombiro}}
\fi
\title{\textbf{Non-intrusive appliance state detection in unseen homes using classical machine learning}}
\author{}
\date{}
\begin{document}
\maketitle
\vspace{-2em}
\ifanon\else
\begin{center}
Zablon Ombiro$^{a,*}$\\[0.5em]
{\small $^{a}$ School of Computing and Engineering Sciences, Strathmore University, Nairobi, Kenya\\
$^{*}$ Corresponding author: Zablon Ombiro, zablon.ombiro@strathmore.edu}
\end{center}
\fi

\section*{Highlights}
\begin{itemize}
\item Features and the appliance, not the model, drive detection of appliance states
\item Appliance explains 38\% of F1 variance, features 13\%, the model only 8\%
\item Pooled testing overstates F1 in unseen homes: 0.548 versus 0.447
\item Feature engineering gives the same 52\% F1 gain in unseen homes
\item House-grouped selection improves F1 for three of five appliances
\end{itemize}

\section*{Graphical abstract}
\begin{figure}[htbp]
\centering
\includegraphics[width=\textwidth]{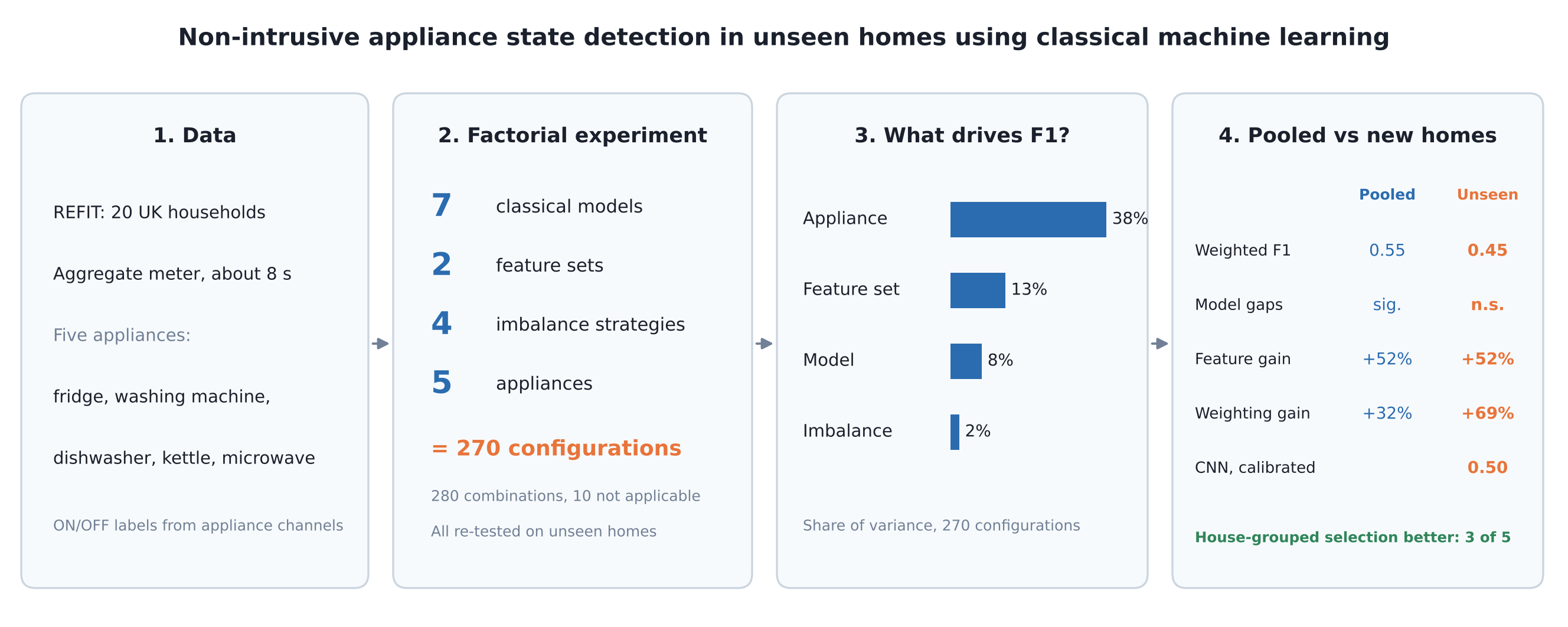}
\caption*{Design and main results of the factorial experiment, including validation on unseen houses.}
\end{figure}

\begin{abstract}
Non-intrusive load monitoring (NILM) infers, from a single smart meter's aggregate signal, when each appliance in a home is ON or OFF, which supports energy feedback and demand response. Classical machine-learning detectors report F1 scores (the harmonic mean of precision and recall) from 0.53 to 0.98, but the causes of this spread are unseparated, so it is unclear which design choices matter. Deployment requires detectors that work in houses never seen during training, whereas several classical studies test on the houses used for training. We propose a factorial analysis of variance design that separates the effects of model, feature set and class-imbalance strategy across five appliances (270 configurations), and a leave-house-out protocol with house-grouped model selection. The design attributes performance to each factor and measures it on unseen homes. On REFIT, a public dataset of 20 UK households, the appliance explains 38\% of F1 variance, the feature set 13\%, the model 8\% and the imbalance strategy 2\%. Pooled testing overstates support-weighted F1 (0.548 versus 0.447 in unseen houses), class weighting gains grow from 32\% to 69\%, and feature engineering gives the same 52\% gain in both protocols. House-grouped selection raises F1 for three of five appliances, by up to 0.110, although the overall gain is small (0.455 versus 0.446). Under the same protocol, a convolutional neural network scores 0.450 (0.496 calibrated) against 0.447 for classical models. The findings guide appliance-level monitoring in smart-metering and energy-management systems.
\end{abstract}

\noindent\textbf{Keywords:} Non-intrusive load monitoring; Appliance state detection; Classical machine learning; Feature engineering; Class imbalance; Unseen-house validation

\section{Introduction}\label{introduction}

Appliance-level monitoring of electricity use is a real-world engineering problem for smart-metering and energy-management systems, because knowing which appliances consume electricity supports energy feedback to occupants, demand response, measurement and verification of efficiency measures, and fault detection in buildings \cite{r1,r2}. Installing a sensor on every appliance is costly and intrusive. Non-intrusive load monitoring (NILM), first formulated by Hart \cite{r3}, instead infers appliance operation from the aggregate signal measured by a single meter. Its simplest actionable form is appliance state detection, also called ON/OFF classification, in which a model decides at each time step whether a given appliance is operating. Because it relies only on the existing meter, NILM can in principle be deployed across installed smart-meter fleets at little additional hardware cost, and classical models are small enough to run on a home energy gateway rather than in a data centre.

Two families of methods dominate NILM classification. Classical machine learning, including linear models, decision trees, instance-based methods, kernel methods and tree ensembles, has been applied to several public datasets \cite{r4,r5,r6,r7}. Deep learning, including convolutional, recurrent and hybrid architectures \cite{r8,r9}, reports the strongest results on the REFIT dataset \cite{r10,r11,r12}: a convolutional recurrent network tested on held-out houses reached a micro-averaged F1 score of 0.63 \cite{r10}. Weighted F1 scores of 0.78 to 0.91 are listed for deep models on REFIT in the comparison of Pirnat et al. \cite{r12}, whose own experiments used an 80:20 split of mixed data synthesised from the measurement datasets. Classical models are simpler and less demanding in computation and memory than deep models \cite{r12}, so whether they can approach deep-learning accuracy determines whether appliance-level monitoring can run on existing low-cost devices.

The published evidence does not settle whether classical models can approach deep-learning accuracy. Reported classical results for ON/OFF classification range from an F1 score of 0.528 with multi-label classifiers on REDD \cite{r4,r13}, as summarised in \cite{r12}, to an F-score of 0.98 with Random Forest on the high-frequency BLUED dataset \cite{r7}. These values are not directly comparable, because the studies differ in dataset, appliance set, features, class-imbalance treatment and evaluation metric. Accuracy is particularly problematic: when an appliance is rarely ON, accuracy is dominated by the OFF class, so a high accuracy can coexist with poor detection of ON events \cite{r14}. A recent meta-analysis of machine-learning NILM studies quantified this inconsistency as near-total heterogeneity (I\textsuperscript{2} $\approx$ 99.995\%) and called for standardised datasets and reproducible evaluation frameworks \cite{r15}.

Consequently, two questions remain open. The first is which design factor drives the performance of classical NILM models. Each previous study fixed its own dataset, model family, feature representation, imbalance treatment and metric, so the literature cannot tell whether a reported advantage comes from the model, the features or the appliance on which it was measured. The second is whether the conclusions of a benchmark survive the way it is validated. Several classical studies train and test on data from the same house \cite{r5,r7}, and training-set sizes are often not reported \cite{r7}. Deep-learning NILM studies routinely test on houses unseen during training \cite{r16,r17,r18}, and grouped validation is recommended when samples are nested within units such as houses \cite{r19}. Leakage between training and test data is a recognised cause of over-optimistic and irreproducible results in applied machine learning \cite{r20}, and in NILM it arises naturally, because samples from the same house share its appliance stock, base load and usage routine. Whether the conclusions of a classical benchmark, namely which model, features and imbalance treatment work best, change under validation on unseen houses has, to our knowledge, not been measured.

To answer these questions, a factorial experiment was run on REFIT \cite{r21}, the dataset used by the three deep-learning references \cite{r10,r11,r12}. The dataset, the appliances, the labels and the protocol were held fixed. The model (seven algorithms from five families), the feature set (basic or engineered) and the class-imbalance strategy (no correction, class weighting, or oversampling by the synthetic minority oversampling technique, SMOTE, or by adaptive synthetic sampling, ADASYN) were varied jointly for five appliances, giving 280 combinations. Of these, 270 could be fitted, because the k-nearest neighbours (k-NN) algorithm has no class-weighting variant, and each was evaluated by nested cross-validation. Four research questions are examined: RQ1, how much of the variation in performance is attributable to each design factor; RQ2, which of the resulting conclusions hold on houses unseen during training; RQ3, how sensitive the headline result is to the training-set size; and RQ4, how far tuned classical models remain from deep learning on the same data. RQ1 is answered in Sections 4.2 to 4.5 and 4.7, RQ2 in Sections 4.8 to 4.10, RQ3 in Section 4.6 and RQ4 in Sections 4.10 and 5.2.

The contributions are as follows. (i) A variance decomposition of the 270 fitted configurations, which shows that the appliance and the feature set account for far more of the performance than the choice of model. (ii) A leave-house-out re-evaluation of all 270 configurations, repeated on four house partitions and complemented by house-grouped hyperparameter tuning, which identifies the conclusions that survive validation on unseen houses and quantifies how pooled validation biases model selection, against imbalance correction. (iii) A house-grouped selection protocol, in which each configuration is chosen by grouped cross-validation on the training houses only, and its nested comparison with pooled selection. (iv) A same-protocol comparison with a convolutional network, which indicates how much of the gap to published deep-learning results depends on the evaluation protocol. (v) A training-set-size sensitivity analysis with a control experiment for a forced change of the hyperparameter grid. (vi) A reporting checklist for NILM benchmarks, and an open release of the code, the verified channel map and all results.

\textbf{Position of this work.} Unlike studies that propose a new detector, this work asks how a classical detector should be designed and whether the design holds in new homes. A factorial design with variance decomposition quantifies how much each design choice matters, a leave-house-out protocol shows that pooled cross-validation gives biased model selection for unseen homes, house-grouped selection reduces that selection loss for three of five appliances, and a same-protocol convolutional network places the classical models in context. For smart-metering and energy-management systems, the practical result is that the appliance and the feature set matter more than the model family, that class weighting is more valuable on unseen homes than pooled validation suggests, and that configurations should be selected with house-grouped validation.

\section{Related work}\label{related-work}

This section reviews classical machine-learning approaches to appliance classification, the deep-learning results reported on REFIT and the evaluation practices that determine how these results can be compared. Broader surveys of NILM methods are available in \cite{r1,r2,r22,r23}.

\subsection{Classical machine learning for appliance classification}\label{classical-machine-learning-for-appliance-classification}

Early classical approaches treated appliance identification as a multi-label problem. Tabatabaei et al. \cite{r4} applied multi-label classifiers, including RAkEL and MLkNN, to the REDD dataset; an average F1 score of 0.528 is attributed to this work in \cite{r12}. Rehmani et al. \cite{r5} compared several classifiers, including k-NN, Random Forest and Extra-Trees, on the low-resolution aggregate power of one REDD house and reported accuracies of up to 98\%. Singh and Majumdar \cite{r6} used multi-label sparse representation-based classification on REDD and Pecan Street data and obtained micro-averaged F1 scores of 0.68 and 0.70, whereas Wu et al. \cite{r7} obtained an F-score of 0.98 with Random Forest and high-frequency electrical features on the BLUED dataset, training and testing on consecutive days of a single house.

More recent studies continue to report results that are specific to a dataset and a metric. Shabbir et al. \cite{r24} compared logistic regression, k-NN and XGBoost with a long short-term memory network on household smart-meter data. Paramasivam Mohan and Sundaram \cite{r25} evaluated decision tree, naive Bayes and k-NN classifiers on REFIT and REDD after balancing the data. Cabral et al. \cite{r26} combined analysis-of-variance feature selection with gradient-boosting machines on REDD, Young et al. \cite{r27} compared k-NN, support vector machines, XGBoost and a convolutional network on PLAID and WHITED, and Alwaz et al. \cite{r28} applied Random Forest, decision tree, k-NN, XGBoost and gradient boosting to forecasting and appliance ON/OFF prediction on hourly PRECON data. None of these studies evaluates the five-appliance REFIT task with a joint variation of models, features and imbalance strategies. Work in this journal approaches deployment from other directions: Yan et al. \cite{r56} detect unknown appliances with normalised k-nearest neighbour distances, Han et al. \cite{r57} identify appliances with topological data analysis, Batic et al. \cite{r58} add attribution priors to the training of load disaggregation models, and Bertalanič and Fortuna \cite{r59} classify appliances from time-series sub-window images. These studies address unseen appliances, new representations and explainability; to our knowledge, none examines whether design conclusions hold in unseen houses, which is the question addressed here.

\subsection{Deep-learning benchmarks on REFIT}\label{deep-learning-benchmarks-on-refit}

Three deep-learning studies provide the reference results for this work, because they use REFIT and the same five appliances. Tanoni et al. \cite{r10} proposed a convolutional recurrent neural network trained with weakly labelled data and, testing on three houses held out from training, reported a micro-averaged F1 score of 0.63. Langevin et al. \cite{r11} proposed a disaggregation model based on variational autoencoders and reported an average F1 score of 0.76 for the ON state. Pirnat et al. \cite{r12} proposed an energy-efficient convolutional transpose recurrent network for multi-label ON/OFF classification of low-frequency data. Under an 80:20 split of mixed data synthesised from the measurement datasets, in windows of about 6 h, it reached a weighted F1 score of 0.91. Their re-run of the model of Tanoni et al. reached 0.83, and their comparison lists 0.78 for the model of Langevin et al. The paper does not state whether the split was random, chronological or by house, so the difference between the 0.63 and 0.83 of the Tanoni architecture cannot be attributed to the split alone, because input data, labels and averaging also differ. Pirnat et al. identified class imbalance as an open issue and named oversampling methods such as ADASYN as future work \cite{r12}. More recent architectures, such as the transformer-based NILMFormer \cite{r29}, continue to advance the state of the art.

\subsection{Class imbalance and evaluation practice}\label{class-imbalance-and-evaluation-practice}

Appliance ON/OFF labels are strongly imbalanced, because most appliances are OFF for most of the day. Under such imbalance, accuracy rewards a classifier that predicts the majority class, and it is systematically less reliable than metrics that focus on the minority class \cite{r14}. The imbalanced-learning literature \cite{r30} offers two main remedies: cost-sensitive learning, such as class weighting, and synthetic oversampling, such as SMOTE \cite{r31} and ADASYN \cite{r32}. No single technique dominates across problems \cite{r33}, the benefit of SMOTE-based methods depends on class overlap and data complexity rather than on the imbalance ratio alone \cite{r34}, and classifier performance does not improve monotonically with rebalancing effort \cite{r35}. Imani et al. \cite{r36} benchmarked Random Forest and XGBoost with SMOTE and ADASYN at controlled imbalance levels on tabular data, but comparable experiments on a physically measured NILM task are scarce.

A related concern is reproducibility: undisclosed training conditions hinder the reproduction of machine-learning results \cite{r37}, and flawed benchmarking protocols can create apparent differences between methods \cite{r38}. For classical NILM models, training-set size is such a condition, because kernel and ensemble methods are often trained on subsamples when the full data are too large.

The validation split is a further condition of this kind. When samples are nested within houses and neighbouring samples are strongly correlated, random splits place samples that share house-specific patterns, and in time series overlapping windows, in both the training and the test sets, which inflates the performance estimate. Blocked or grouped cross-validation is recommended for such data \cite{r19}, and leakage of this kind is a documented source of irreproducible results across scientific applications of machine learning \cite{r20}. In NILM, evaluation on houses unseen during training has been common for deep models since Kelly and Knottenbelt \cite{r16}. The lack of standardised evaluation has been identified as a barrier to comparability \cite{r39}, and reviews of NILM evaluation have catalogued the resulting diversity of metrics and protocols \cite{r40,r41}. Open toolkits such as NILMTK \cite{r42} standardise data handling and metrics, but the choice of validation split remains with the researcher. In the classical studies reviewed above, however, the split is either pooled or not stated, and its effect on which model, features or imbalance strategy is found best has, to our knowledge, not been quantified.

\subsection{Research gap}\label{research-gap}

Table 1 compares the reviewed studies with the present work. Existing studies either evaluate classical models on a single dataset with a single evaluation choice, or evaluate deep models on REFIT without examining class-imbalance handling. Specifically, there is a lack of studies that:

\begin{itemize}
\item
  vary the model family, the feature representation and the imbalance strategy jointly under a single protocol and quantify the share of performance attributable to each;
\item
  test whether the conclusions of such a benchmark hold on houses unseen during training; and
\item
  report the training-set size of each model and test the sensitivity of the results to it.
\end{itemize}

The present work meets all three conditions on REFIT, with the appliances used by the deep-learning benchmarks, and compares its results with those references. It therefore proposes a controlled factorial design with variance decomposition for classical models, a leave-house-out protocol that tests whether benchmark conclusions transfer to unseen homes, and a house-grouped model-selection protocol, compared in a nested design with pooled selection, all validated on public data with public code.

{\footnotesize\setlength{\tabcolsep}{3pt}
\begin{longtable}{>{\raggedright\arraybackslash}p{0.0489\dimexpr\linewidth-24pt\relax}>{\raggedright\arraybackslash}p{0.3456\dimexpr\linewidth-24pt\relax}>{\raggedright\arraybackslash}p{0.2918\dimexpr\linewidth-24pt\relax}>{\raggedright\arraybackslash}p{0.3136\dimexpr\linewidth-24pt\relax}}
\caption{Comparative analysis of related work and the present contribution.}\label{tab:1}\\
\toprule
\textbf{Ref.} & \textbf{Approach and data} & \textbf{Limitation} & \textbf{Contribution of this work}\tabularnewline
\midrule
\cite{r4} & Multi-label classifiers (RAkEL, MLkNN); REDD & Single dataset; no feature or imbalance ablation & Features and imbalance strategies varied jointly under one protocol\tabularnewline
\cite{r5} & k-NN, Random Forest, Extra-Trees and other classifiers; one REDD house & Accuracy under strong imbalance; not comparable with F1 & ON-class F1 as primary metric; accuracy bias quantified\tabularnewline
\cite{r6} & Multi-label sparse representation classification; REDD, Pecan Street & Single model family and feature representation & Seven models from five algorithm families\tabularnewline
\cite{r7} & Random Forest with high-frequency electrical features; BLUED (one house) & Single house, temporal split; training-set size not reported & Training-set size reported and its effect tested\tabularnewline
\cite{r24,r25,r26,r27,r28} & LR, DT, k-NN, SVM and gradient boosting; smart-meter data, REFIT, REDD, PLAID, WHITED, PRECON & Each study fixes its model set, features and dataset & Joint sweep on REFIT with the appliances of the deep-learning benchmarks\tabularnewline
\cite{r10,r11,r12} & Recurrent, convolutional and autoencoder-based deep models; REFIT & High computational cost; class-imbalance handling not examined & Classical reference on the same dataset, including four imbalance strategies\tabularnewline
\bottomrule
\end{longtable}
}

\section{Materials and methods}\label{materials-and-methods}

The experimental framework is summarised in Fig. 1. Aggregate and appliance power were taken from REFIT, ON/OFF labels were derived from the appliance channels, two feature sets were computed from the aggregate signal, and every combination of model and imbalance strategy was evaluated by nested cross-validation. The following subsections describe each step.

\begin{figure}[htbp]
\centering
\includegraphics[width=1.000\textwidth]{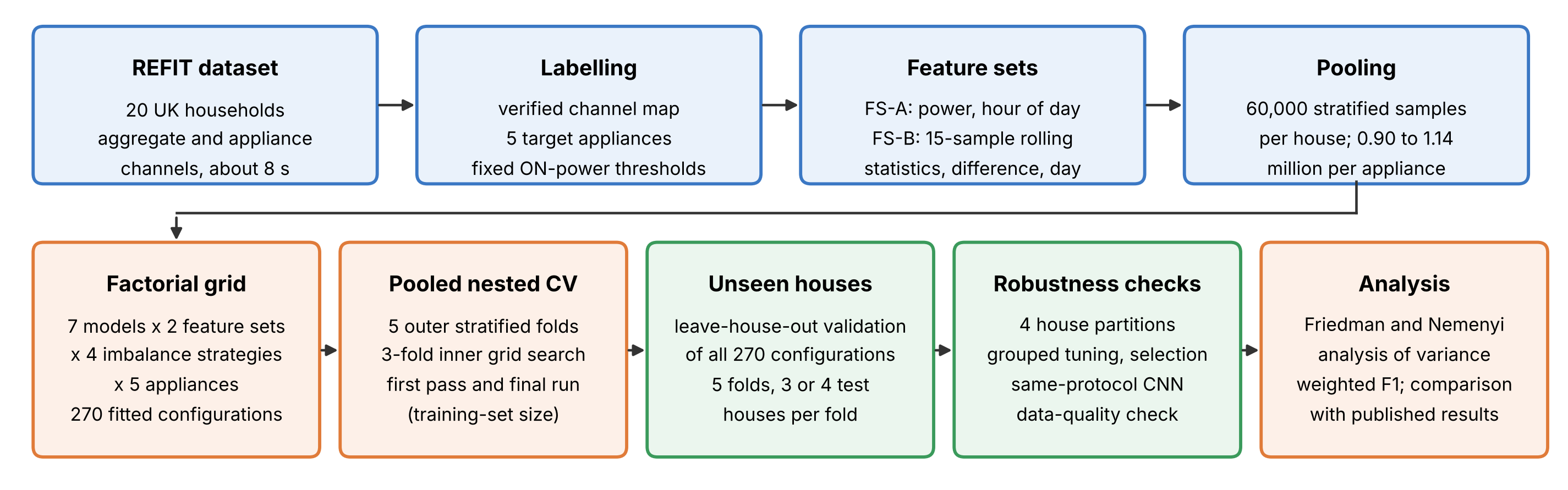}
\caption{Experimental framework of the study.}\label{fig:1}
\end{figure}

\subsection{Dataset}\label{dataset}

REFIT \cite{r21} contains electricity measurements from 20 households in the United Kingdom recorded over approximately two years. Each house provides the aggregate power and up to nine individually metered appliance channels at a resolution of approximately 8 s, and each house contributes between 4.4 and 7.4 million samples. The dataset is openly available under a CC BY 4.0 licence. It was selected for three reasons: it is the dataset used by the three deep-learning references \cite{r10,r11,r12}, so the results refer to the same data; it is openly accessible without registration; and its size supports a large, pooled training set for each appliance. The houses are numbered 1 to 13 and 15 to 21. Because the assignment of appliances to channels differs between houses, a per-house channel map was compiled from the dataset documentation and checked against the measured data before labelling.

\subsection{Target appliances and ground-truth labels}\label{target-appliances-and-ground-truth-labels}

Five appliances were selected: fridge, washing machine, dishwasher, kettle and microwave. They are the appliances used in the REFIT benchmarks \cite{r10,r12} and they include a continuously cycling load, two programme-based loads and two short high-power loads. ON/OFF labels were derived from each appliance channel with the fixed ON-power thresholds of Kelly and Knottenbelt \cite{r16}, without their minimum ON and OFF durations: a sample is labelled ON when the appliance power reaches the threshold given in Table 2. The same thresholds were applied to all houses.

{\footnotesize\setlength{\tabcolsep}{3pt}
\begin{longtable}{>{\raggedright\arraybackslash}p{0.3154\dimexpr\linewidth-36pt\relax}>{\raggedright\arraybackslash}p{0.1369\dimexpr\linewidth-36pt\relax}>{\raggedright\arraybackslash}p{0.1369\dimexpr\linewidth-36pt\relax}>{\raggedright\arraybackslash}p{0.1369\dimexpr\linewidth-36pt\relax}>{\raggedright\arraybackslash}p{0.1369\dimexpr\linewidth-36pt\relax}>{\raggedright\arraybackslash}p{0.1369\dimexpr\linewidth-36pt\relax}}
\caption{Target appliances, coverage and class balance. Median values are taken over the houses that monitor each appliance; the per-house audit is given in Supplementary Table S1.}\label{tab:2}\\
\toprule
\textbf{Appliance} & \textbf{Houses} & \textbf{Median ON time (\%)} & \textbf{ON share in pooled data (\%)} & \textbf{Median ON power (W)} & \textbf{ON threshold (W)}\tabularnewline
\midrule
Fridge & 18 & 33.6 & 31.5 & 94 & 50\tabularnewline
Washing machine & 19 & 2.46 & 3.14 & 551 & 20\tabularnewline
Dishwasher & 15 & 2.08 & 4.73 & 836 & 10\tabularnewline
Kettle & 16 & 0.71 & 0.62 & 2729 & 2000\tabularnewline
Microwave & 16 & 0.27 & 0.38 & 1281 & 200\tabularnewline
\bottomrule
\end{longtable}
}

Table 2 shows that class imbalance is severe and differs strongly between appliances. The fridge is ON for a median of 33.6\% of the time, whereas the kettle and the microwave are ON for less than 1\% of the time. This imbalance motivates the class-imbalance strategies described in Section 3.5.

Several data-quality issues were identified from the per-house audit (Supplementary Table S1) and retained in the main analysis. The aggregate signals of Houses 11 and 21 are affected by rooftop solar generation. The fridge channel of House 11 is ON for only 0.22\% of the time at a mean ON power of approximately 1007 W, against a median of 33.6\% and 53 to 117 W in the other houses, so it does not behave as a fridge. The kettle channel of House 4 is ON for only 0.004\% of the time, the dishwasher channel of House 21 is ON for 13.1\% of the time at 356 W, against a median of 2.1\%, and the kettle channels of Houses 17 and 21 share a monitor with a toaster. A sensitivity analysis without the solar-affected houses and the House 4 kettle channel is reported in Section 4.10.

A separate, correctable problem was found in the per-house channel map. A channel-by-channel check against the dataset documentation and the measured data revealed five incorrect or missing assignments, all for the kettle and the microwave: the kettle channel was wrongly mapped in House 12 and missing in Houses 15 and 21, and the microwave channel was wrongly mapped in Houses 12 and 13. The map was corrected, the kettle and microwave feature tables were rebuilt, all kettle and microwave experiments were repeated, and a verification script is included with the code. All results reported here use the corrected map, so the kettle models were trained and evaluated on all 16 houses that monitor a kettle. Where a house monitors two candidate channels, the choice is documented with the code; in House 4, the more frequently used of two washing machines was selected.

\subsection{Data preparation and feature sets}\label{data-preparation-and-feature-sets}

For each appliance, the aggregate and appliance signals of every house that monitors the appliance were loaded at full resolution. Features were computed at full resolution before any subsampling, because rolling-window features require temporal continuity. Each house was then reduced to 60,000 samples by stratified random sampling, which preserves the ON/OFF ratio of the house, and the houses were pooled into a single dataset per appliance. The pooled datasets contain between 900,000 samples (dishwasher, 15 houses) and 1,140,000 samples (washing machine, 19 houses). The per-house limit keeps training tractable and prevents houses with longer records from dominating the pooled data. The house identifier was retained for traceability but was not used as a feature.

Two feature sets were constructed from the aggregate signal (Table 3). Feature set A (FS-A) contains only the instantaneous aggregate power and the hour of day, and represents a minimal input set. Feature set B (FS-B) adds rolling statistics of the kind widely used in time-series classification \cite{r43}, which describe the local shape of the signal: the rolling mean, standard deviation, minimum, maximum, skewness and kurtosis of the aggregate power, its first difference, the rolling mean of the absolute first difference, and the day of week. The rolling window covers the 15 most recent samples, approximately 2 min at the REFIT sampling rate, so the features use no future samples; its length was not tuned for individual appliances.

{\footnotesize\setlength{\tabcolsep}{3pt}
\begin{longtable}{>{\raggedright\arraybackslash}p{0.4213\dimexpr\linewidth-24pt\relax}>{\raggedright\arraybackslash}p{0.4223\dimexpr\linewidth-24pt\relax}>{\raggedright\arraybackslash}p{0.0782\dimexpr\linewidth-24pt\relax}>{\raggedright\arraybackslash}p{0.0782\dimexpr\linewidth-24pt\relax}}
\caption{Feature sets. P\textsubscript{t} denotes the aggregate active power at time step t; rolling statistics are computed over a window of 15 samples.}\label{tab:3}\\
\toprule
\textbf{Feature} & \textbf{Definition} & \textbf{FS-A} & \textbf{FS-B}\tabularnewline
\midrule
Aggregate power & P\textsubscript{t} (W) & Yes & Yes\tabularnewline
Hour of day & 0 to 23 & Yes & Yes\tabularnewline
Day of week & 0 to 6 & No & Yes\tabularnewline
Rolling mean and standard deviation & Mean and standard deviation of P & No & Yes\tabularnewline
Rolling minimum and maximum & Minimum and maximum of P & No & Yes\tabularnewline
Rolling skewness and kurtosis & Third and fourth standardised moments of P & No & Yes\tabularnewline
First difference & $\Delta$P\textsubscript{t} = P\textsubscript{t} $-$ P\textsubscript{t$-$1} & No & Yes\tabularnewline
Rolling mean absolute difference & Mean of \textbar $\Delta$P\textbar{} & No & Yes\tabularnewline
Number of features & & 2 & 11\tabularnewline
\bottomrule
\end{longtable}
}

\subsection{Classification models}\label{classification-models}

Seven models were selected to cover the algorithm families used in the classical NILM literature (Table 4): logistic regression (LR) as a linear baseline; a decision tree (DT) as a single-tree baseline; k-NN as an instance-based method; a support vector machine with a radial basis function kernel (SVM-RBF) \cite{r44} as a kernel method; and three tree ensembles, Random Forest (RF) \cite{r45}, XGBoost \cite{r46} and LightGBM \cite{r47}. Tree ensembles are the family behind several of the strongest classical NILM results \cite{r5,r7}. Inputs were standardised for LR, k-NN and SVM-RBF. Hyperparameters were tuned by grid search within the nested cross-validation (Section 3.6), so that every model received a comparable tuning effort. A multilayer perceptron was not included, because it is a neural network and would blur the distinction between classical and deep models on which the study is based.

{\footnotesize\setlength{\tabcolsep}{3pt}
\begin{longtable}{>{\raggedright\arraybackslash}p{0.0928\dimexpr\linewidth-18pt\relax}>{\raggedright\arraybackslash}p{0.1838\dimexpr\linewidth-18pt\relax}>{\raggedright\arraybackslash}p{0.7234\dimexpr\linewidth-18pt\relax}}
\caption{Classification models and hyperparameter grids. All models used a random seed of 42.}\label{tab:4}\\
\toprule
\textbf{Model} & \textbf{Family} & \textbf{Hyperparameter grid}\tabularnewline
\midrule
LR & Linear & Inverse regularisation strength C $\in$ \{0.1, 1, 10\}; standardised inputs\tabularnewline
DT & Single tree & Maximum depth $\in$ \{6, 12, unlimited\}; minimum samples per leaf $\in$ \{1, 5\}\tabularnewline
k-NN & Instance-based & Number of neighbours k $\in$ \{3, 5, 11\}; standardised inputs\tabularnewline
SVM-RBF & Kernel & C $\in$ \{1, 10\}; kernel coefficient $\gamma$ $\in$ \{scale, auto\}; standardised inputs\tabularnewline
RF & Bagging ensemble & Number of trees $\in$ \{100, 300\}; maximum depth $\in$ \{12, unlimited\} in the first pass and \{12\} in the final run\tabularnewline
XGBoost & Boosting ensemble & Number of trees $\in$ \{100, 300\}; maximum depth $\in$ \{4, 8\}\tabularnewline
LightGBM & Boosting ensemble & Number of trees $\in$ \{100, 300\}; maximum depth $\in$ \{8, unlimited\}\tabularnewline
\bottomrule
\end{longtable}
}

\subsection{Class-imbalance handling}\label{class-imbalance-handling}

Four strategies were compared for every combination of appliance, feature set and model. The first applies no correction. The second, class weighting (CW), uses the native cost-sensitive mechanism of each model: balanced class weights for LR, DT, SVM-RBF, RF and LightGBM, and a positive-class weight equal to the ratio of OFF to ON samples in each training fold for XGBoost. The third and fourth are the synthetic oversampling methods SMOTE \cite{r31} and ADASYN \cite{r32}. Oversampling was applied only to the training data within each cross-validation split, so that no synthetic sample reached a validation or test fold. Because k-NN has no native class-weighting mechanism, the combination of k-NN and CW was recorded as not applicable rather than approximated. SMOTE was included as the standard synthetic oversampling method \cite{r31}, and ADASYN because it was named as future work for REFIT by Pirnat et al. \cite{r12}.

\subsection{Evaluation protocol and metrics}\label{evaluation-protocol-and-metrics}

Each configuration was evaluated by nested cross-validation (CV), which avoids the optimistic bias of tuning and testing on the same folds \cite{r48,r49}. The outer loop used five stratified folds, shuffled with a fixed seed, to estimate generalisation performance. Within each outer training fold, a three-fold grid search selected the hyperparameters that maximised the F1 score. The folds were formed from the pooled samples, so training and test folds contain samples from the same houses; a complementary validation on unseen houses is described in Section 3.9. The nested design removes the bias of tuning hyperparameters on the test folds, but the best of the 54 configurations of an appliance is still chosen on the outer folds, so the best pooled scores are slightly optimistic; this selection effect is examined on unseen houses in Section 4.9.

Performance was measured by the F1 score of the ON class, the harmonic mean of precision \emph{P} and recall \emph{R}:

\begin{equation}\label{eq:1}
P = \frac{\text{TP}}{TP + FP},\quad R = \frac{\text{TP}}{TP + FN},\quad F1 = \frac{2PR}{P + R}
\end{equation}

where \emph{TP}, \emph{FP} and \emph{FN} are the numbers of true positive, false positive and false negative samples for the ON class. Precision, recall and accuracy were recorded for every outer fold, and all scores are reported as means over the five folds. To compare with the deep-learning benchmarks, which report a weighted F1 score over appliance labels, the per-appliance scores were combined into a support-weighted F1 score:

\begin{equation}\label{eq:2}
{F1}_{w} = \frac{\sum_{a = 1}^{5}{n_{a}{F1}_{a}}}{\sum_{a = 1}^{5}n_{a}}
\end{equation}

where \emph{n}\textsubscript{a} is the number of ON samples of appliance \emph{a} and F1\textsubscript{a} is its F1 score. Accuracy was recorded but not used to rank configurations, because it is dominated by the OFF class under the imbalance shown in Table 2 \cite{r14}.

\subsection{Training-set size and final run}\label{training-set-size-and-final-run}

The training cost of SVM-RBF and the tree ensembles grows rapidly with the number of samples. An initial trial showed that the full grid would take several days on the available computer, an Apple MacBook Pro (M3 Pro processor, 18 GB of memory; Supplementary Table S16), so in a first pass these four models were trained on a fixed random subsample of each pooled dataset (Table 5). LR, DT and k-NN were always trained on the full pooled data. Every capped configuration was flagged in the results log with its original and capped sample counts.

Because the cap was imposed for computational reasons only, the four capped models were re-trained without it, and these re-run results are the final results of this study. RF, XGBoost and LightGBM completed the re-run on the full pooled data. For RF, the unlimited-depth option was removed from the grid after repeated out-of-memory failures on datasets of up to 1,140,000 samples, so the maximum depth was fixed at 12; the effect of this restriction is quantified in Section 5.3. The uncapped SVM-RBF run completed none of its 40 configurations in approximately 9 h, so the kernel method does not scale to the full REFIT datasets within a practical budget. SVM-RBF was therefore re-run with a larger cap of 100,000 samples drawn by stratified sampling, which preserves the ON/OFF ratio of the rare-event appliances.

The final result of each configuration is thus the re-run value for RF, XGBoost, LightGBM and SVM-RBF and the original value for LR, DT and k-NN. The 160 matched configurations of the four re-run models provide a direct measure of the effect of training-set size (Section 4.6).

{\footnotesize\setlength{\tabcolsep}{3pt}
\begin{longtable}{>{\raggedright\arraybackslash}p{0.1307\dimexpr\linewidth-18pt\relax}>{\raggedright\arraybackslash}p{0.4149\dimexpr\linewidth-18pt\relax}>{\raggedright\arraybackslash}p{0.4544\dimexpr\linewidth-18pt\relax}}
\caption{Training-set size per model in the first pass and in the final run.}\label{tab:5}\\
\toprule
\textbf{Model} & \textbf{First pass} & \textbf{Final run}\tabularnewline
\midrule
LR, DT, k-NN & Full pooled data & Full pooled data (not re-run)\tabularnewline
SVM-RBF & 30,000 samples, uniform random & 100,000 samples, stratified\tabularnewline
RF & 60,000 samples, uniform random & Full pooled data; maximum depth fixed at 12\tabularnewline
XGBoost & 100,000 samples, uniform random & Full pooled data\tabularnewline
LightGBM & 100,000 samples, uniform random & Full pooled data\tabularnewline
\bottomrule
\end{longtable}
}

\subsection{Statistical analysis}\label{statistical-analysis}

Differences between models were tested with the Friedman test on the best F1 score of each model for each appliance, with the five appliances as blocks, followed by the Nemenyi post-hoc test \cite{r50}. For the main comparison, the Iman-Davenport F statistic recommended for few blocks is reported alongside the $\chi$\textsuperscript{2} statistic. k-NN, which has no class-weighting variant, contributes its best of six rather than eight configurations per appliance. Two models differ significantly at level $\alpha$ when their average ranks differ by more than the critical difference

\begin{equation}\label{eq:3}
CD = q_{\alpha}\sqrt{\frac{k(k + 1)}{6N}}
\end{equation}

where \emph{k} is the number of models, \emph{N} is the number of appliances and \emph{q}\textsubscript{$\alpha$} is the critical value of the Studentised range statistic divided by $\sqrt{2}$ \cite{r50}. With \emph{k} = 7, \emph{N} = 5 and $\alpha$ = 0.05, CD = 4.03. Because five outer folds are too few for a paired significance test between individual configurations, differences between the best configurations of an appliance were assessed against the fold-to-fold standard deviation. No correction for multiple comparisons was applied across analyses, and configurations share data and models, so p-values are reported as descriptive rather than confirmatory.

\subsection{Validation on unseen houses}\label{validation-on-unseen-houses}

Because the folds of the nested cross-validation contain samples from the same houses, and samples from the same house share its appliance stock, base load and usage routine, the pooled scores may overstate performance in a new home. All 270 fitted configurations were therefore re-evaluated with leave-house-out validation. The houses of each appliance were divided into five groups by stratified group k-fold splitting, so that each fold was tested on three or four houses not seen in training. The hyperparameters of each configuration were fixed at the values most often selected in its pooled nested cross-validation, oversampling was applied to the training houses only, SVM-RBF was trained on a stratified subsample of 100,000 samples and all other models on the full training data. A first run on this reference partition re-evaluated the best engineered-feature configuration of each model for each appliance (35 configurations; Section 4.8), and the full factorial then covered all 270 configurations (Section 4.9). Four robustness checks followed (Section 4.10). First, the factorial of the four fastest models (LR, DT, XGBoost and LightGBM; 160 configurations) was repeated on three further random house partitions. Second, these four models were re-tuned with engineered features by a three-fold grouped grid search over the training houses only, and the inner grouped scores also chose the configuration of each fold, which defines the house-grouped selection protocol: tune and choose within the training houses by grouped folds, then evaluate on the held-out houses. Comparing it with pooled selection gives a nested comparison. Third, a one-dimensional convolutional network (CNN), similar in design to sequence-to-point models \cite{r8}, was trained on the same houses, partition, labels and metric. Its input was a window of 99 samples (about 13 min) of the aggregate power and its first difference, centred on each of 60,000 stratified target samples per house; it had four convolutional layers and was trained for six epochs with a class-weighted loss, without tuning and with a decision threshold of 0.5. A calibrated variant held out about 20\% of the training houses for early stopping (at most 12 epochs, with a patience of three) and threshold selection, and was run with the centred window and with a causal window of the 50 samples up to the target sample. Fourth, the four fastest models and the fridge configuration were repeated without the solar-affected Houses 11 and 21 and the House 4 kettle channel. In all experiments, the metrics were computed on the test houses of each fold and averaged over the five folds, the support-weighted F1 score was computed with Eq. (2), and models were compared with the Friedman test as in Section 3.8. In total, more than 5,000 models were fitted and tested on held-out houses.

\subsection{Variance decomposition and selection analysis}\label{variance-decomposition-and-selection-analysis}

To quantify how much each design factor contributes to the reported performance (RQ1), the F1 scores of the 270 fitted configurations were analysed by a fixed-effects analysis of variance with the appliance, the model, the feature set and the imbalance strategy as factors, first with main effects only and then with all two-way interactions. The share of variance of each term was its type II sum of squares divided by the sum of the type II sums of squares of all terms and the residual, which approximates the $\eta^{2}$ effect size \cite{r51} in this unbalanced design. Because k-NN has no class-weighting variant, the model with interactions was fitted to the 240 configurations of the six other models; the main-effects model uses all 270. The residuals were not normal (Shapiro-Wilk p = 0.003), their variance differed between appliances (Levene p \textless{} 0.001) and the configurations share data, so the p-values are descriptive and a rank-based analysis was run as a check. To compare the two protocols on identical configurations (RQ2), the main-effects decomposition was applied separately to the pooled and leave-house-out scores of all 270 configurations. For the 35-configuration experiment, the fold-level F1 scores were decomposed into the appliance, the group of test houses nested within the appliance, and the model, using sequential sums of squares in that order. The over-optimism of a configuration was its pooled F1 score minus its leave-house-out F1 score. For each appliance, the configuration with the highest pooled F1 score was compared with the one with the highest leave-house-out F1 score, which is an upper bound because it is chosen on the test houses; the nested house-grouped selection of Section 3.9 provides the unbiased comparison. The computing environment and run time of each experiment are reported in Supplementary Table S16.

\section{Results}\label{results}

Of the 280 configurations, 270 were fitted; the remaining 10 correspond to the combination of k-NN and CW, which is not applicable. Unless stated otherwise, the results in this section are the final results defined in Section 3.7. The complete results of both passes are given in Supplementary Tables S2 to S4.

\subsection{Accuracy and F1 under class imbalance}\label{accuracy-and-f1-under-class-imbalance}

Accuracy exceeded the F1 score in all 270 configurations. The mean accuracy was 0.862 (range 0.496 to 0.996), whereas the mean F1 score was 0.278 (range 0.000 to 0.570). The divergence is largest for the rarest appliance. For the microwave with FS-A and no correction, XGBoost and SVM-RBF did not detect a single ON sample in any outer fold (F1 = 0.000), yet each achieved an accuracy of 0.996. Judged by accuracy, these models would appear almost perfect, although they learned nothing about the ON class. This result confirms on the present data why accuracy is unsuitable for comparing NILM classifiers under class imbalance \cite{r14}, and it explains why the accuracy reported by Rehmani et al. \cite{r5} cannot be compared with F1 scores (Section 5.2).

\subsection{Model comparison}\label{model-comparison}

Fig. 2 shows the best F1 score of each model on each appliance, averaged over the five appliances. XGBoost achieved the highest mean (0.498 $\pm$ 0.099), followed by DT (0.458 $\pm$ 0.117), LightGBM (0.450 $\pm$ 0.151), k-NN (0.445 $\pm$ 0.121) and RF (0.437 $\pm$ 0.167). SVM-RBF (0.406 $\pm$ 0.156) and LR (0.264 $\pm$ 0.143) were clearly weaker. The large standard deviations reflect differences between appliances rather than instability of the models, since the fold-to-fold standard deviation of the best configurations did not exceed 0.040 (Section 4.5).

\begin{figure}[htbp]
\centering
\includegraphics[width=1.000\textwidth]{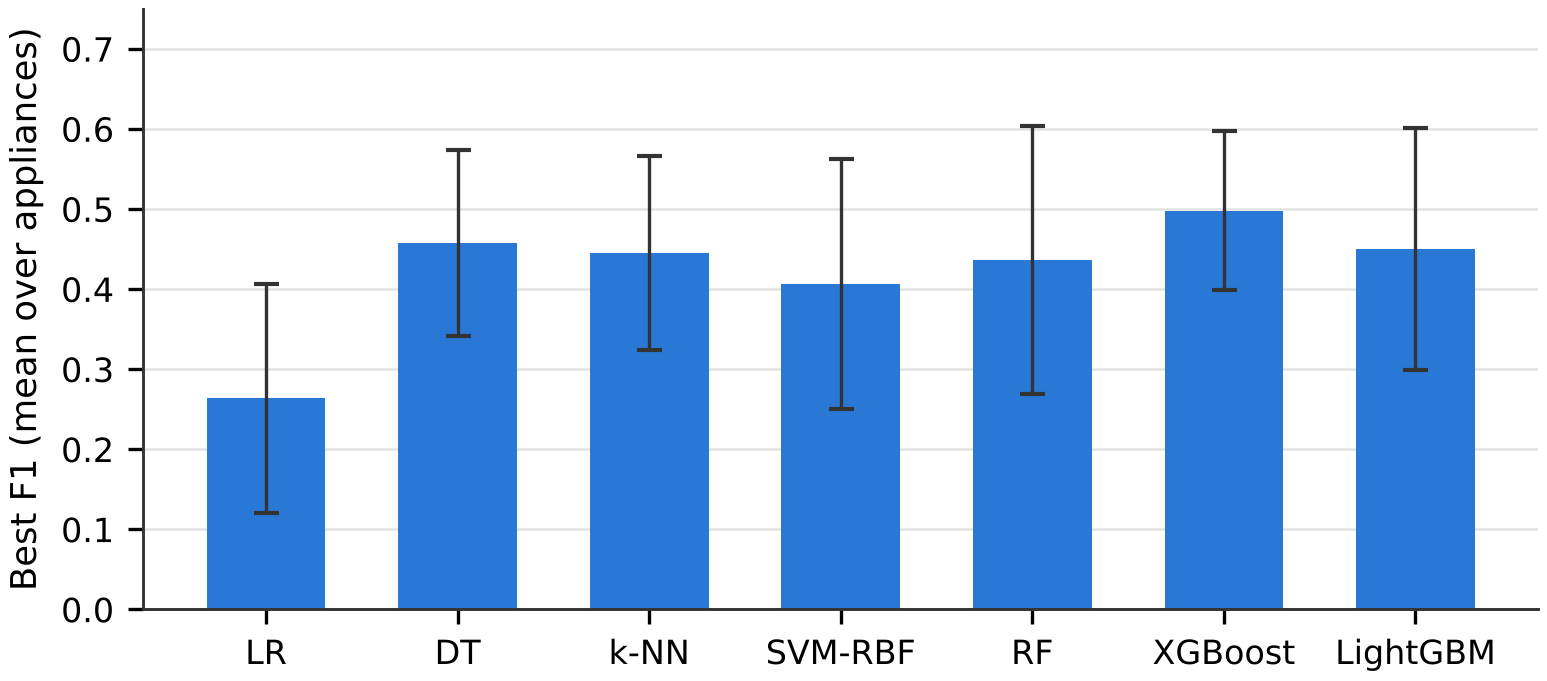}
\caption{Best F1 score of each model, averaged over the five appliances (best feature set and imbalance strategy per appliance). Error bars show the standard deviation over the appliances.}\label{fig:2}
\end{figure}

The Friedman test rejected the hypothesis that all seven models perform equally ($\chi$\textsuperscript{2} = 21.17, p = 0.002; Iman-Davenport F(6, 24) = 9.59, p < 0.001). The average ranks and the Nemenyi critical difference are shown in Fig. 3. XGBoost ranked first for four appliances and second for the fridge, giving an average rank of 1.2. The rank differences between XGBoost and LR (5.8) and between XGBoost and SVM-RBF (4.2) exceed the critical difference of 4.03 and are significant at $\alpha$ = 0.05. The other differences, including between DT and LR (3.8), are not significant, which reflects the limited power of a test based on five appliances. Model choice therefore matters in aggregate, and XGBoost is the most consistent model, but the data do not support a strict ranking among the tree-based models and k-NN.

\begin{figure}[htbp]
\centering
\includegraphics[width=1.000\textwidth]{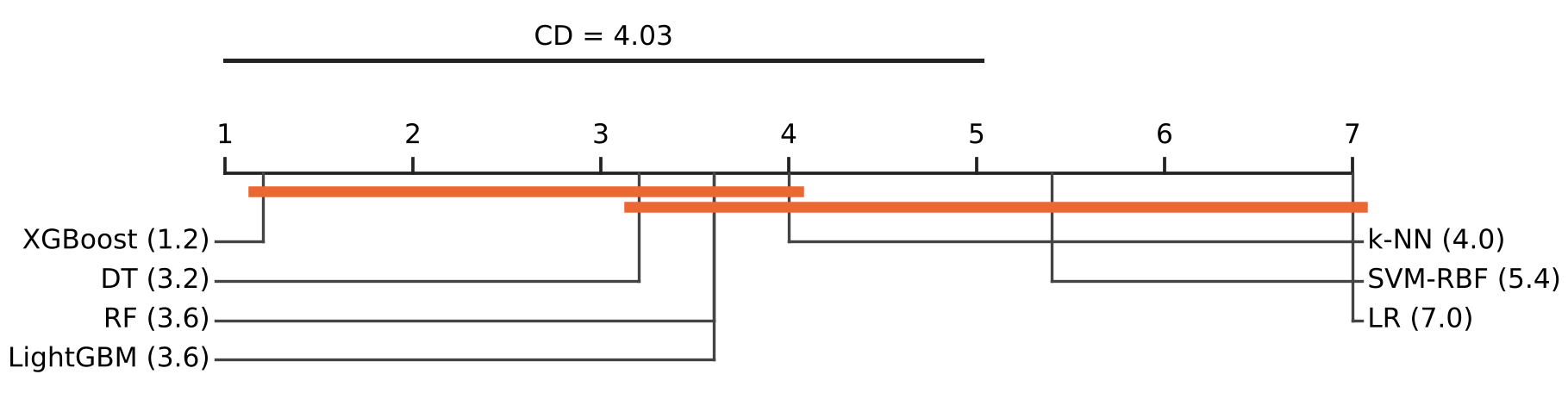}
\caption{Critical-difference diagram of the Nemenyi test ($\alpha$ = 0.05). Models connected by a horizontal bar do not differ significantly.}\label{fig:3}
\end{figure}

\subsection{Effect of feature engineering}\label{effect-of-feature-engineering}

Fig. 4 compares the two feature sets using the best F1 score of each model. At this best-case level the effect of FS-B is mostly positive but uneven: DT, RF, XGBoost and LightGBM gained between 0.019 and 0.043, SVM-RBF gained 0.040 and k-NN gained 0.140, while LR was slightly better with FS-A (by 0.034). The picture changes when all configurations are considered. Averaged over every model, imbalance strategy and appliance, the F1 score rose from 0.220 with FS-A to 0.335 with FS-B, a relative improvement of 52\%. This is the largest average effect of any factor in the study.

\begin{figure}[htbp]
\centering
\includegraphics[width=1.000\textwidth]{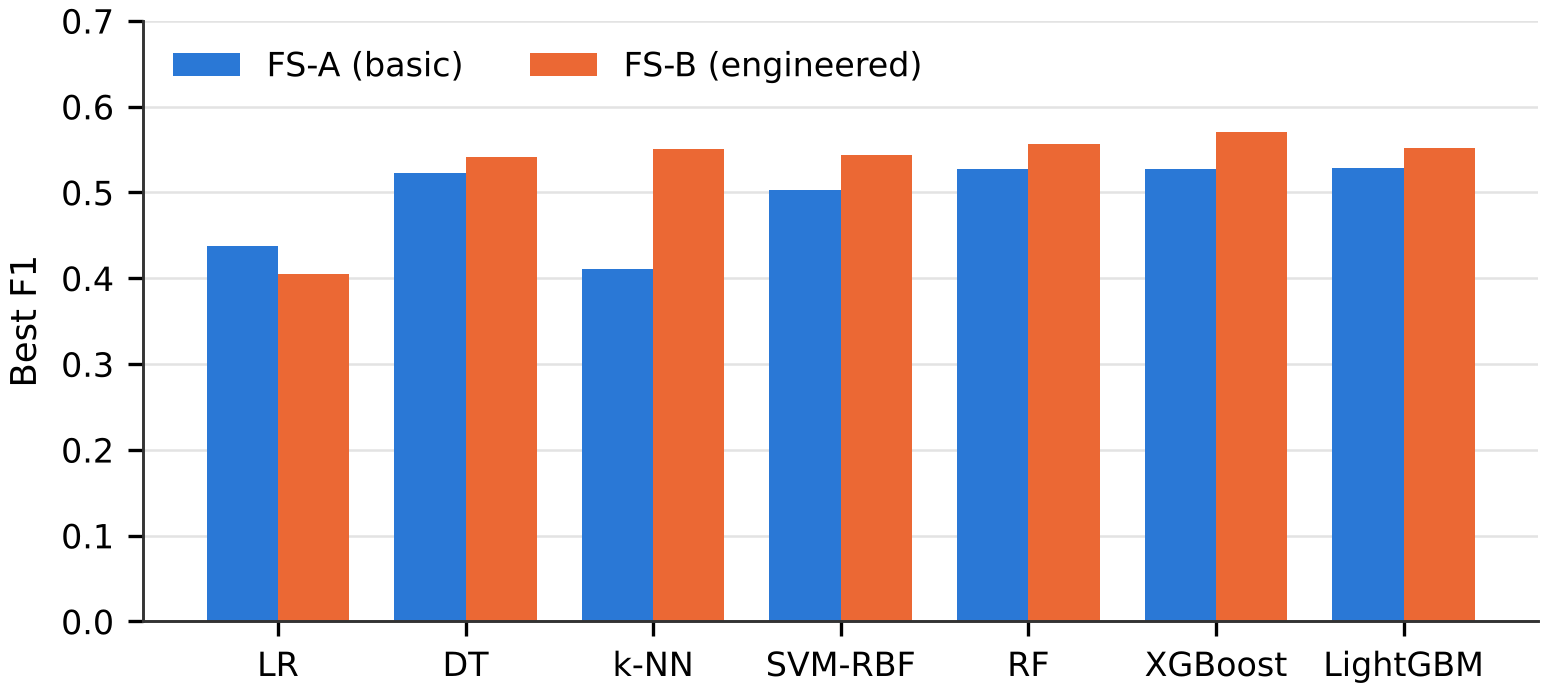}
\caption{Best F1 score of each model with the basic (FS-A) and engineered (FS-B) feature sets, maximised over appliances and imbalance strategies.}\label{fig:4}
\end{figure}

The small gains in Fig. 4 reflect the maximum over appliances, which is set by the appliance on which each model performs best. When the best F1 score of each model is averaged over the five appliances instead, FS-B raised it by 0.136 to 0.231 for every model except LR (0.014); for XGBoost, for example, from 0.267 to 0.498. The rolling statistics of FS-B supply information about the shape of the load that the instantaneous reading does not contain, and all five best configurations use FS-B (Section 4.5).

\subsection{Effect of class-imbalance handling}\label{effect-of-class-imbalance-handling}

Fig. 5 compares the four imbalance strategies using the best F1 score of each model. Averaged over the six models that support all four strategies (k-NN has no class weighting), CW produced the highest best-case F1 score (0.518), followed by ADASYN (0.509), SMOTE (0.501) and no correction (0.467). Over all configurations of these six models, the mean F1 score was 0.305 with CW and 0.231 without correction, a relative improvement of 32\%, with SMOTE (0.291) and ADASYN (0.274) in between. The largest gain was obtained for LR, whose best F1 score rose from 0.109 without correction to 0.438 with ADASYN.

\begin{figure}[htbp]
\centering
\includegraphics[width=1.000\textwidth]{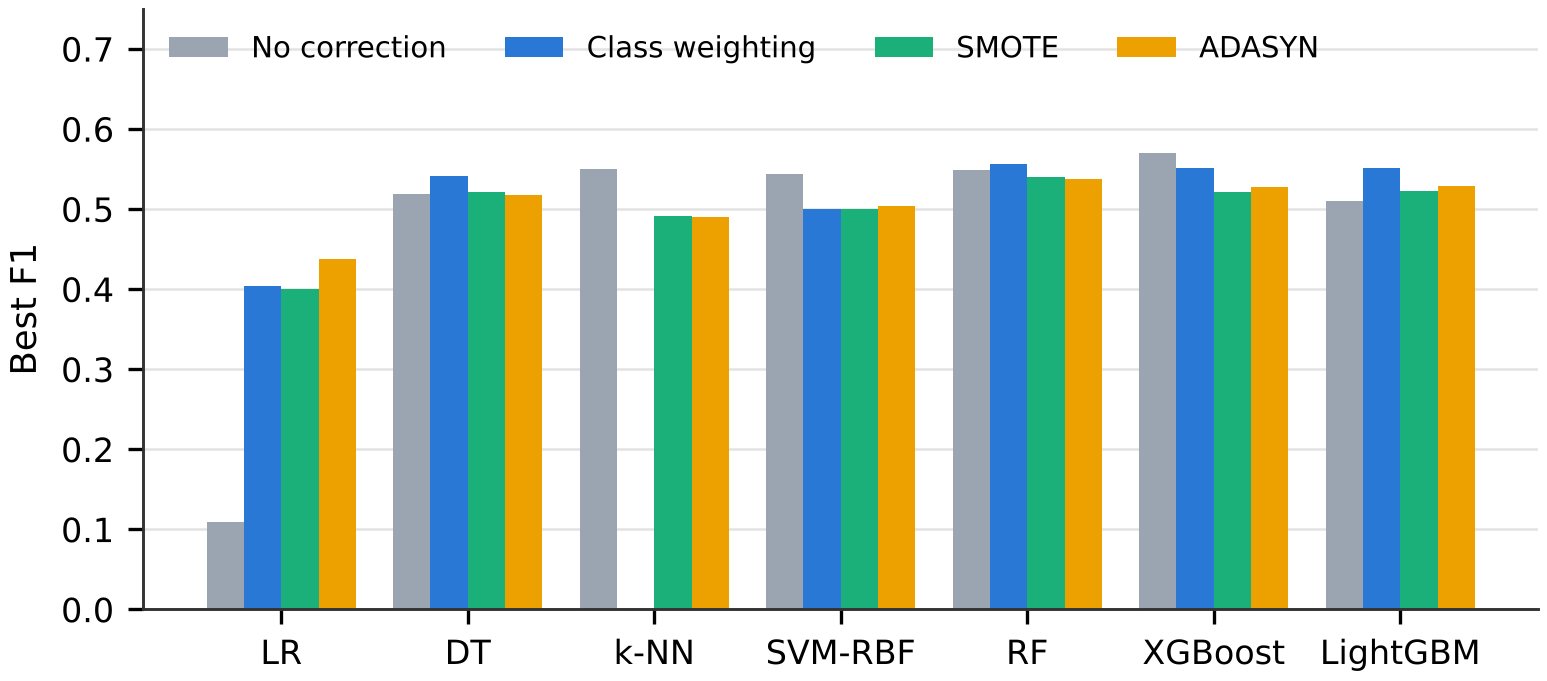}
\caption{Best F1 score of each model for each imbalance strategy, maximised over appliances and feature sets. Class weighting is not applicable to k-NN.}\label{fig:5}
\end{figure}

The average conceals a strong dependence on the appliance. Table 6 reports the mean F1 score of the four tree-based models with FS-B for each strategy. CW improved the fridge from 0.330 to 0.550 and the microwave from 0.202 to 0.228, but it reduced the F1 score of the washing machine (from 0.487 to 0.428), the dishwasher (from 0.500 to 0.344) and the kettle (from 0.537 to 0.412). Oversampling was not the best strategy for any appliance.

{\footnotesize\setlength{\tabcolsep}{3pt}
\begin{longtable}{>{\raggedright\arraybackslash}p{0.3655\dimexpr\linewidth-30pt\relax}>{\raggedright\arraybackslash}p{0.1586\dimexpr\linewidth-30pt\relax}>{\raggedright\arraybackslash}p{0.1586\dimexpr\linewidth-30pt\relax}>{\raggedright\arraybackslash}p{0.1586\dimexpr\linewidth-30pt\relax}>{\raggedright\arraybackslash}p{0.1586\dimexpr\linewidth-30pt\relax}}
\caption{Mean F1 score of the tree-based models (DT, RF, XGBoost and LightGBM) with FS-B for each imbalance strategy. The best value for each appliance is shown in bold.}\label{tab:6}\\
\toprule
\textbf{Appliance} & \textbf{No correction} & \textbf{Class weighting} & \textbf{SMOTE} & \textbf{ADASYN}\tabularnewline
\midrule
Fridge & 0.330 & \textbf{0.550} & 0.504 & 0.502\tabularnewline
Washing machine & \textbf{0.487} & 0.428 & 0.435 & 0.409\tabularnewline
Dishwasher & \textbf{0.500} & 0.344 & 0.336 & 0.313\tabularnewline
Kettle & \textbf{0.537} & 0.412 & 0.334 & 0.299\tabularnewline
Microwave & 0.202 & \textbf{0.228} & 0.198 & 0.194\tabularnewline
\bottomrule
\end{longtable}
}

The benefit of CW does not follow the degree of imbalance. The fridge, the least imbalanced appliance, gained most, whereas the washing machine, the dishwasher and the kettle, which are ON for less than 3\% of the time, lost performance. Reweighting the ON class raises the number of predicted ON events, which increases recall at the cost of precision; when the engineered features already separate the ON events well, this trade-off lowers the F1 score. The result agrees with evidence that the benefit of rebalancing depends on class overlap and data complexity rather than on the imbalance ratio \cite{r34,r35}. Consequently, two studies that each apply a different imbalance strategy to a different appliance can reach very different conclusions, which contributes to the heterogeneity of the literature.

\subsection{Per-appliance performance}\label{per-appliance-performance}

Fig. 6 shows the best F1 score of every model on every appliance, and Table 7 lists the best configuration for each appliance. The best F1 score ranged from 0.324 for the microwave (XGBoost, FS-B, CW) to 0.570 for the kettle (XGBoost, FS-B, no correction), which is the highest of the five appliances. The fridge (0.556), the washing machine (0.532) and the dishwasher (0.512) were in between. For the fridge, RF, XGBoost and LightGBM were within 0.005 of each other. The fold-to-fold standard deviation of the best configurations was between 0.002 and 0.040, which is small compared with the differences between appliances.

\begin{figure}[htbp]
\centering
\includegraphics[width=1.000\textwidth]{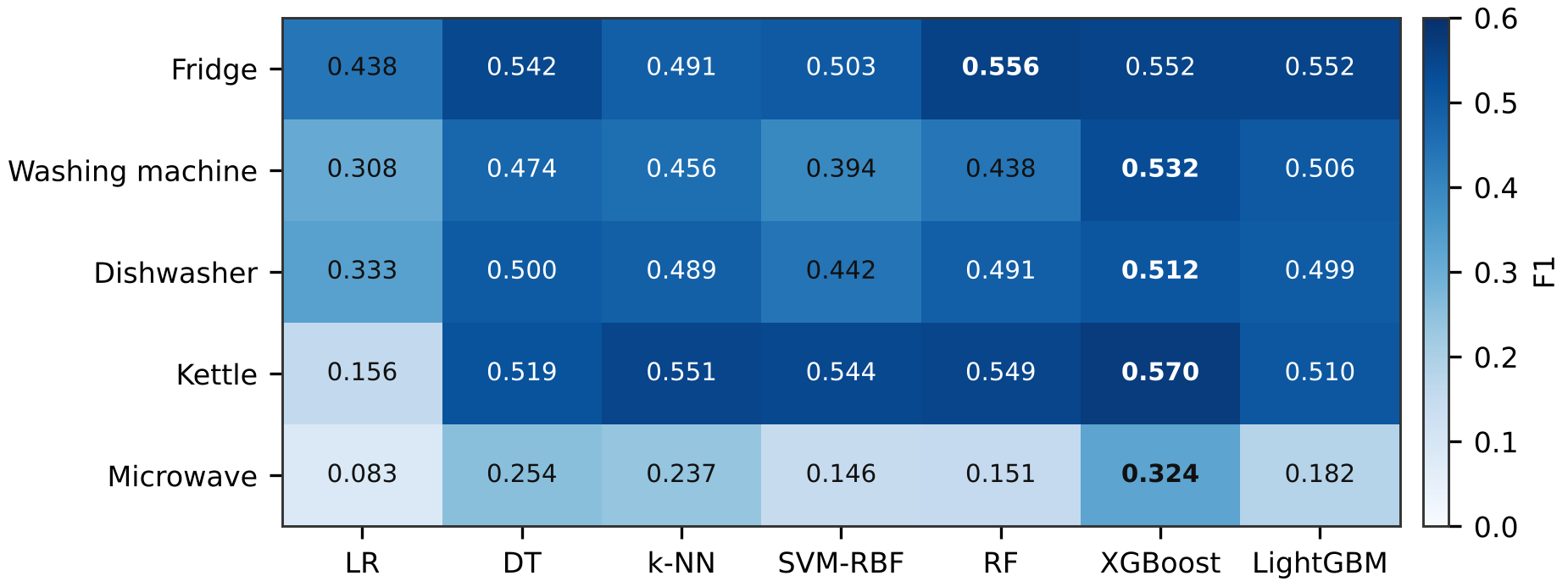}
\caption{Best F1 score of each model on each appliance (final results). The best model for each appliance is shown in bold.}\label{fig:6}
\end{figure}

{\footnotesize\setlength{\tabcolsep}{3pt}
\begin{longtable}{>{\raggedright\arraybackslash}p{0.1912\dimexpr\linewidth-48pt\relax}>{\raggedright\arraybackslash}p{0.2761\dimexpr\linewidth-48pt\relax}>{\raggedright\arraybackslash}p{0.0817\dimexpr\linewidth-48pt\relax}>{\raggedright\arraybackslash}p{0.0817\dimexpr\linewidth-48pt\relax}>{\raggedright\arraybackslash}p{0.1029\dimexpr\linewidth-48pt\relax}>{\raggedright\arraybackslash}p{0.0817\dimexpr\linewidth-48pt\relax}>{\raggedright\arraybackslash}p{0.0817\dimexpr\linewidth-48pt\relax}>{\raggedright\arraybackslash}p{0.1029\dimexpr\linewidth-48pt\relax}}
\caption{Best configuration for each appliance (final results). F1 is the mean over the five outer folds and SD its standard deviation; the last row is the support-weighted F1 score of Eq. (2).}\label{tab:7}\\
\toprule
\textbf{Appliance} & \textbf{Configuration} & \textbf{F1} & \textbf{SD} & \textbf{Precision} & \textbf{Recall} & \textbf{Accuracy} & \textbf{ON samples}\tabularnewline
\midrule
Fridge & RF, FS-B, CW & 0.556 & 0.002 & 0.444 & 0.746 & 0.626 & 339,857\tabularnewline
Washing machine & XGBoost, FS-B, none & 0.532 & 0.006 & 0.690 & 0.433 & 0.976 & 35,826\tabularnewline
Dishwasher & XGBoost, FS-B, none & 0.512 & 0.008 & 0.790 & 0.378 & 0.966 & 42,545\tabularnewline
Kettle & XGBoost, FS-B, none & 0.570 & 0.021 & 0.779 & 0.450 & 0.996 & 5,979\tabularnewline
Microwave & XGBoost, FS-B, CW & 0.324 & 0.040 & 0.276 & 0.444 & 0.992 & 3,632\tabularnewline
Weighted F1 & & 0.548 & & & & & 427,839\tabularnewline
\bottomrule
\end{longtable}
}

The support-weighted F1 score over the five appliances is 0.548, and the unweighted mean is 0.499. The weighted value is dominated by the fridge, which accounts for 79\% of the ON samples. The differences between the best and second-best configurations were small for three appliances, 0.005 for the fridge, 0.012 for the dishwasher and 0.020 for the kettle, compared with 0.026 for the washing machine and 0.044 for the microwave. Near-tied configurations should therefore be regarded as practically equivalent.

The operating points of the best configurations (Table 7) show that no appliance combines high precision with high recall. The fridge configuration detects 74.6\% of the ON samples but also labels 43.0\% of the OFF samples as ON, which gives a precision of 0.444. The washing machine, the dishwasher and the kettle show the opposite behaviour: precision is between 0.690 and 0.790, but only 37.8\% to 45.0\% of the ON samples are detected. The microwave is weak on both measures (precision 0.276, recall 0.444), which reflects the small number of ON samples available for learning (0.38\% of the pooled data).

\subsection{Sensitivity to training-set size}\label{sensitivity-to-training-set-size}

Changing the training-set size reduced the mean F1 score of the 160 re-run configurations (Supplementary Table S4) by 0.021, and the mean accuracy by 0.010. Nevertheless, the best configuration changed for four of the five appliances (Table 8), although for the fridge, the dishwasher and the kettle the two candidates differ by 0.020 or less, which Section 4.5 treats as practically equivalent. The best model changed for three appliances: LightGBM was replaced by RF for the fridge, DT by XGBoost for the dishwasher, and k-NN by XGBoost for the kettle. For the washing machine, XGBoost remained the best model but the best imbalance strategy changed from CW to no correction. For the microwave, the best model (XGBoost) and strategy (CW) were unchanged, yet its best F1 score rose from 0.256 to 0.324, the largest gain of any appliance, which shows that training-set size alone can materially change a benchmark's headline number.

{\footnotesize\setlength{\tabcolsep}{3pt}
\begin{longtable}{>{\raggedright\arraybackslash}p{0.1844\dimexpr\linewidth-36pt\relax}>{\raggedright\arraybackslash}p{0.2727\dimexpr\linewidth-36pt\relax}>{\raggedright\arraybackslash}p{0.0800\dimexpr\linewidth-36pt\relax}>{\raggedright\arraybackslash}p{0.2868\dimexpr\linewidth-36pt\relax}>{\raggedright\arraybackslash}p{0.0800\dimexpr\linewidth-36pt\relax}>{\raggedright\arraybackslash}p{0.0960\dimexpr\linewidth-36pt\relax}}
\caption{Best configuration for each appliance in the first pass and in the final run (model, feature set, imbalance strategy).}\label{tab:8}\\
\toprule
\textbf{Appliance} & \textbf{First pass} & \textbf{F1} & \textbf{Final run} & \textbf{F1} & \textbf{Change}\tabularnewline
\midrule
Fridge & LightGBM, FS-B, CW & 0.548 & RF, FS-B, CW & 0.556 & +0.008\tabularnewline
Washing machine & XGBoost, FS-B, CW & 0.512 & XGBoost, FS-B, none & 0.532 & +0.021\tabularnewline
Dishwasher & DT, FS-B, none & 0.500 & XGBoost, FS-B, none & 0.512 & +0.012\tabularnewline
Kettle & k-NN, FS-B, none & 0.551 & XGBoost, FS-B, none & 0.570 & +0.020\tabularnewline
Microwave & XGBoost, FS-B, CW & 0.256 & XGBoost, FS-B, CW & 0.324 & +0.069\tabularnewline
\bottomrule
\end{longtable}
}

The effect of the cap also differed between models (Fig. 7). SVM-RBF improved slightly on average, by 0.002, with 28 of its 40 configurations improving. XGBoost declined marginally, by 0.005, with 18 of 40 configurations improving. LightGBM and RF declined more, by 0.027 and 0.053. The decline of RF is consistent with the change of its search space in the final run, in which the maximum depth was fixed at 12 (Section 3.7); Section 5.3 shows directly that most of this decline is attributable to the depth restriction itself rather than to the larger training set. The model ranking changed accordingly: in the first pass XGBoost already ranked first (average rank 1.6, against 2.8 for DT), and in the final results its lead widened (1.2). The Friedman test on the four re-run models remained significant ($\chi$\textsuperscript{2} = 10.20, p = 0.017).

\begin{figure}[htbp]
\centering
\includegraphics[width=1.000\textwidth]{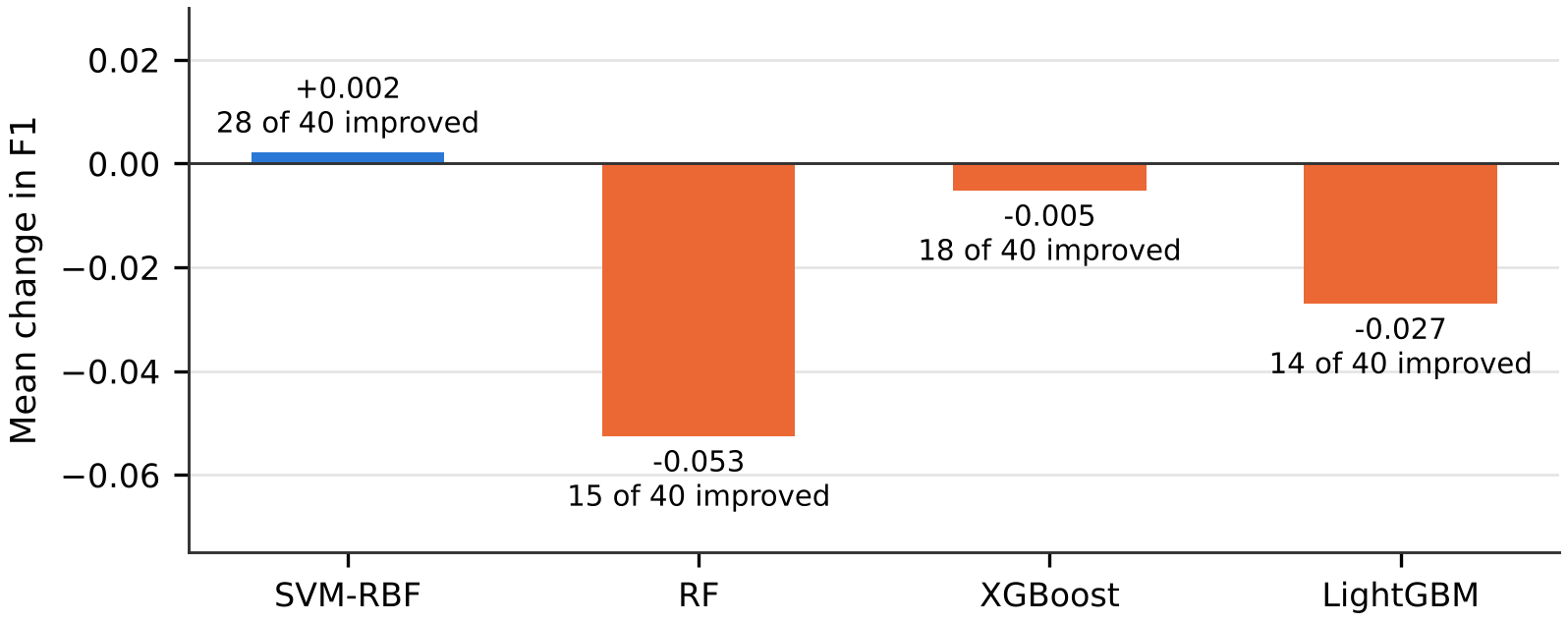}
\caption{Mean change in F1 score between the first pass and the final run for each re-run model, with the number of improved configurations.}\label{fig:7}
\end{figure}

\subsection{Relative importance of the design factors}\label{relative-importance-of-the-design-factors}

Fig. 8a decomposes the variance of the F1 score. In the main-effects model over the 270 fitted configurations, the appliance accounted for 37.7\% of the variance, the feature set for 13.2\%, the model for 7.5\% and the imbalance strategy for 2.1\%, all significant (p \textless{} 0.005); a rank-based analysis gave similar shares (39.5\%, 12.9\%, 7.7\% and 2.3\%). Adding the two-way interactions, fitted to the 240 configurations of the six models with all four strategies, raised the explained variance to 93.9\%. The largest interaction was between the appliance and the imbalance strategy (14.0\%), which is as large as the main effect of the feature set in this model (14.3\%): imbalance handling changes performance strongly, but in opposite directions for different appliances (Table 6), so its average effect is small. The interaction between the appliance and the model was the only one that was not significant (1.1\%, p = 0.07), so the ranking of the models depends little on the appliance. Overall, the appliance accounts for more variance than the main effects of all three modelling choices combined (37.7\% against 22.8\%), although the modelling choices regain weight through their interactions with the appliance, and among the modelling choices the feature representation matters more than the algorithm.

\begin{figure}[htbp]
\centering
\includegraphics[width=1.000\textwidth]{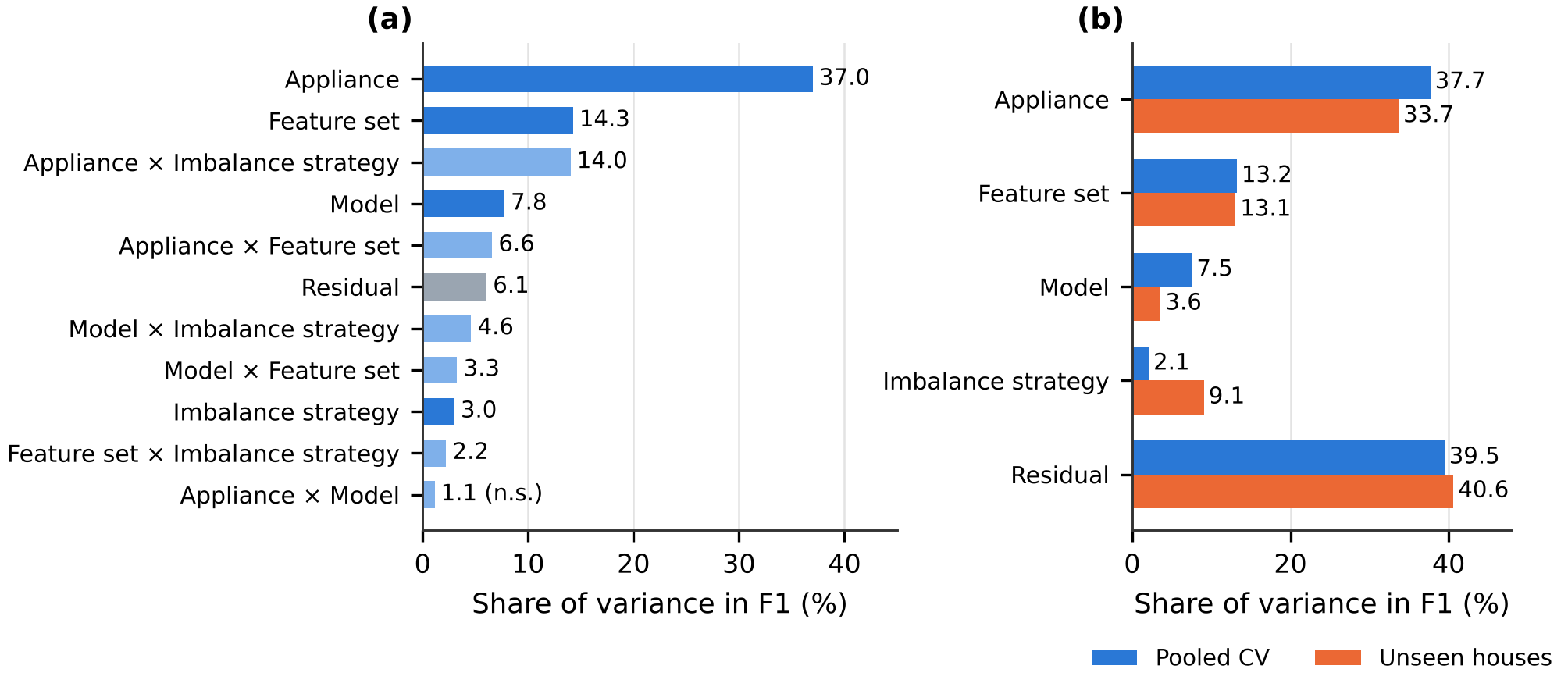}
\caption{Share of the variance in F1 attributed to each design factor. (a) Main effects and two-way interactions under pooled cross-validation, fitted to the 240 configurations of the six models that support all four imbalance strategies (n.s., not significant at $\alpha$ = 0.05; the main-effects shares over all 270 configurations are given in the text). (b) Main effects for all 270 configurations under pooled cross-validation and on unseen houses.}\label{fig:8}
\end{figure}

\subsection{Validation on unseen houses: best configurations and models}\label{validation-on-unseen-houses-best-configurations-and-models}

Table 9 compares the pooled cross-validation scores of the best configurations with their leave-house-out scores. The F1 score of every appliance was lower on unseen houses, and the support-weighted F1 score fell from 0.548 to 0.447, while the unweighted mean fell from 0.499 to 0.363. The decrease was 0.058 for the washing machine, 0.082 for the fridge, 0.100 for the microwave and 0.149 for the kettle, and it was largest for the dishwasher (0.288), whose recall fell from 0.378 to 0.156. This suggests that the models learned household-specific usage patterns that do not transfer, which is plausible for a programme-based load whose timing depends on the household. The decrease is not an artefact of averaging F1 over folds that contain different houses: when the confusion counts of the five folds are pooled before the F1 score is computed, the support-weighted F1 score on unseen houses is 0.457.

{\footnotesize\setlength{\tabcolsep}{3pt}
\begin{longtable}{>{\raggedright\arraybackslash}p{0.2116\dimexpr\linewidth-48pt\relax}>{\raggedright\arraybackslash}p{0.0904\dimexpr\linewidth-48pt\relax}>{\raggedright\arraybackslash}p{0.0904\dimexpr\linewidth-48pt\relax}>{\raggedright\arraybackslash}p{0.0904\dimexpr\linewidth-48pt\relax}>{\raggedright\arraybackslash}p{0.2297\dimexpr\linewidth-48pt\relax}>{\raggedright\arraybackslash}p{0.0995\dimexpr\linewidth-48pt\relax}>{\raggedright\arraybackslash}p{0.0904\dimexpr\linewidth-48pt\relax}>{\raggedright\arraybackslash}p{0.0976\dimexpr\linewidth-48pt\relax}}
\caption{Pooled cross-validation and leave-house-out results for the best configuration of each appliance (Table 7). SD is the standard deviation of F1 over the five leave-house-out folds, and the range is the lowest and highest fold. Changes are computed from unrounded values.}\label{tab:9}\\
\toprule
\textbf{Appliance} & \textbf{F1, pooled} & \textbf{F1, unseen houses} & \textbf{SD} & \textbf{Range over folds} & \textbf{Precision} & \textbf{Recall} & \textbf{Change}\tabularnewline
\midrule
Fridge & 0.556 & 0.474 & 0.112 & 0.292 to 0.580 & 0.380 & 0.692 & $-$0.082\tabularnewline
Washing machine & 0.532 & 0.474 & 0.086 & 0.330 to 0.537 & 0.619 & 0.392 & $-$0.058\tabularnewline
Dishwasher & 0.512 & 0.224 & 0.082 & 0.099 to 0.295 & 0.457 & 0.156 & $-$0.288\tabularnewline
Kettle & 0.570 & 0.421 & 0.160 & 0.229 to 0.591 & 0.511 & 0.366 & $-$0.149\tabularnewline
Microwave & 0.324 & 0.224 & 0.069 & 0.126 to 0.307 & 0.199 & 0.279 & $-$0.100\tabularnewline
Weighted F1 & 0.548 & 0.447 & & & & & $-$0.101\tabularnewline
\bottomrule
\end{longtable}
}

The order of the appliances also changed. Under pooled cross-validation the kettle had the highest F1 score (0.570), but on unseen houses the fridge and the washing machine were highest (0.474 each) and the kettle fell to 0.421, so part of the kettle's advantage in Section 4.5 came from house-specific patterns. The scores varied more between folds on unseen houses (SD between 0.069 and 0.160) than between the pooled folds (at most 0.040), because each fold tests a different set of households, and the ranges in Table 9 show that some households are much harder than others. The support-weighted F1 score of 0.447 is between 0.33 and 0.46 below the deep-learning benchmarks (Section 5.2).

Across the seven models (Supplementary Table S6), the configurations selected by pooled validation lost between 0.007 (LR) and 0.135 (XGBoost) on average. The ranking of the models remained similar to the pooled ranking (Spearman correlation of mean ranks 0.88, p = 0.009), with XGBoost again first (mean rank 3.0, against 1.2 under pooled validation) and LR last (5.8), but the differences between models were no longer significant (Friedman $\chi^{2}$ = 4.80, p = 0.57). For the dishwasher, LR with class weighting scored 0.302 on unseen houses because it lost little (0.027), whereas the other models lost between 0.19 (SVM-RBF) and 0.30 (RF).

The 35 configurations show two further patterns. First, the configurations with the highest pooled F1 scores lost the most on unseen houses (Spearman $\rho$ = 0.53, p = 0.001). Because the pooled score enters both sides of this correlation, the association is weaker when the loss is related to the mean of the two scores ($\rho$ = 0.26, p = 0.13), so with 35 configurations it is not conclusive. Second, the variance of the fold-level F1 scores on unseen houses was dominated by the appliance (52.8\%) and by the group of test houses within each appliance (27.3\%), whereas the model accounted for 4.6\% (sequential sums of squares). The houses on which a model is tested therefore account for about six times more of this variance than the choice among the seven models.

\subsection{Validation on unseen houses: feature sets and imbalance strategies}\label{validation-on-unseen-houses-feature-sets-and-imbalance-strategies}

Fig. 9 compares the pooled and unseen-house F1 scores of all 270 configurations. The two scores were strongly correlated (Spearman $\rho$ = 0.91), and 207 of the 270 configurations scored lower on unseen houses, but the losses were not uniform. Configurations with higher pooled scores lost more on unseen houses ($\rho$ = 0.63, p \textless{} 0.001); because the pooled score appears on both sides of this correlation, part of the association is regression to the mean, but it remains when the loss is related to the mean of the two scores ($\rho$ = 0.51, p \textless{} 0.001).

\begin{figure}[htbp]
\centering
\includegraphics[width=0.700\textwidth]{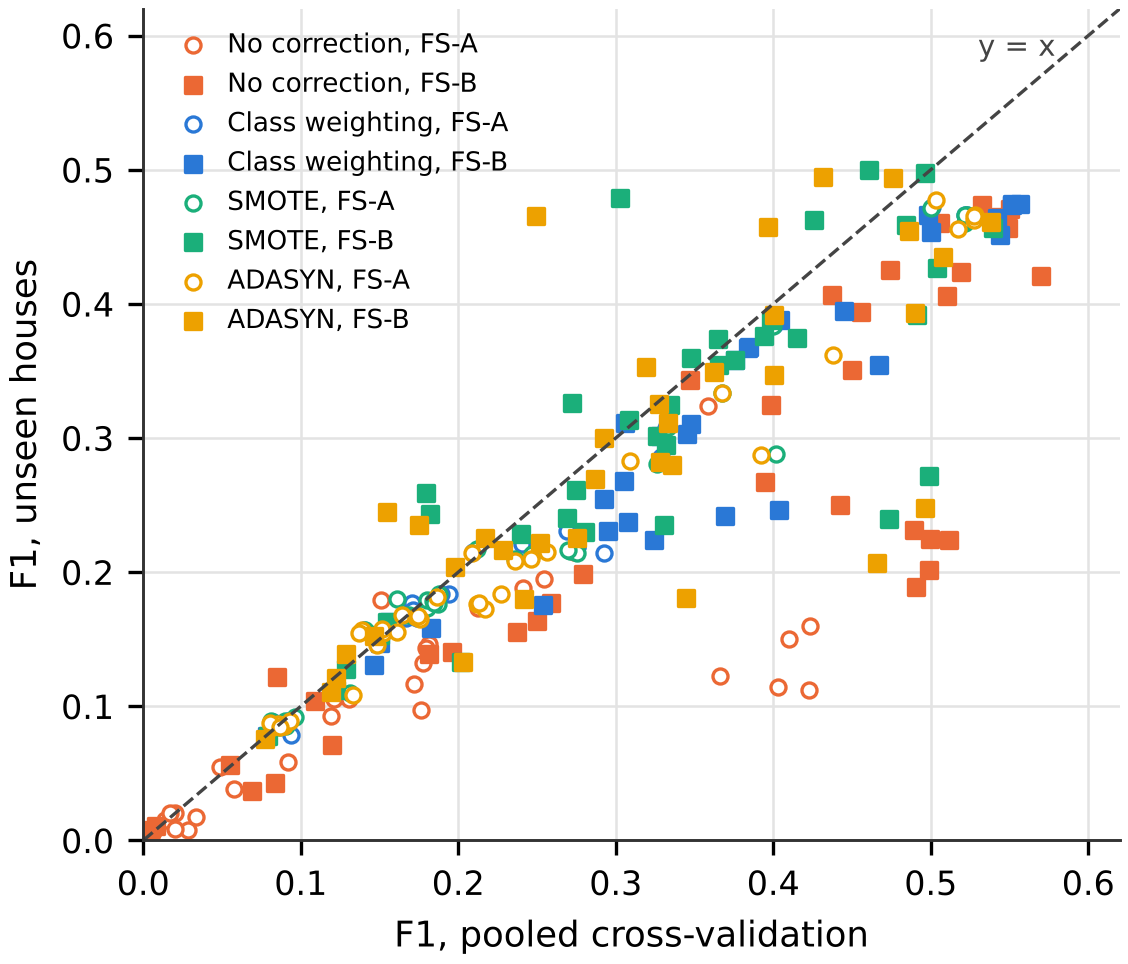}
\caption{Pooled and unseen-house F1 scores of the 270 configurations. Open markers, FS-A; filled markers, FS-B. Points below the dashed line scored lower on unseen houses.}\label{fig:9}
\end{figure}

Feature engineering remained the strongest modelling lever. Averaged over the seven models, four strategies and five appliances, FS-B raised the F1 score from 0.189 to 0.288 on unseen houses, a relative gain of 52\%, the same as under pooled validation (52\%, from 0.220 to 0.335), and FS-B was better than FS-A in 112 of 135 matched configurations (Wilcoxon signed-rank test, p \textless{} 0.001). The engineered features were, however, more over-optimistic than the basic features (mean loss 0.047 against 0.031). Overlapping rolling windows may contribute to this difference, but the basic features, which use no window, also lost performance, so most of the optimism reflects differences between houses rather than near-duplicate samples. For the fridge, their advantage disappeared on unseen houses (change $-$0.019, against +0.026 under pooled validation).

Imbalance handling behaved differently. Among the six models that support all four strategies, configurations without correction lost the most on unseen houses (mean loss 0.072, against 0.021 to 0.036 for the three corrections), so the relative benefit of class weighting over no correction rose from 32\% under pooled validation to 69\% on unseen houses. The variance decomposition shows the same shift (Fig. 8b): the share of variance attributable to the imbalance strategy rose from 2.1\% to 9.1\%, whereas the share attributable to the model fell from 7.5\% to 3.6\%; the shares of the appliance (37.7\% and 33.7\%) and of the feature set (13.2\% and 13.1\%) changed little. A plausible explanation is a shift in class priors and conditional distributions between households \cite{r52,r53}: the prevalence and timing of appliance use differ between households, so an uncorrected model learns the base rate and the habits of the training houses, which helps within those houses and fails in new ones.

As a result, the configuration with the highest pooled score was not the best on unseen houses for any of the five appliances (Table 10). In four of the five cases the best unseen-house configuration used SMOTE, and LightGBM replaced XGBoost for three appliances. The F1 score lost by the pooled selection was 0.150 for the dishwasher, 0.058 for the kettle, 0.026 for the washing machine, 0.019 for the microwave and 0.003 for the fridge (negligible). Because the best configuration on unseen houses is chosen on the test houses, this comparison is an upper bound on what better selection can achieve. In the nested comparison, each fold's configuration was chosen by house-grouped cross-validation on the training houses only, among 16 engineered-feature configurations of LR, DT, XGBoost and LightGBM per appliance (Supplementary Table S13). The effect was smaller but remained: house-grouped selection improved the unseen-house F1 score by 0.110 for the dishwasher, 0.035 for the washing machine and 0.030 for the microwave, and was within 0.007 of the pooled selection for the fridge and the kettle. It chose an imbalance correction in 22 of the 25 folds, against 10 of 25 for the pooled selection, and its support-weighted F1 score was 0.455, against 0.446 for the pooled selection among the same configurations and 0.471 for an oracle that picks the best of these configurations on the test houses of each fold. Fold by fold, house-grouped selection was better in 12 of 25 folds and worse in 10 (Wilcoxon signed-rank test over the 25 folds, p = 0.09), so the overall gain is small, is not statistically significant and is concentrated in three appliances. The model ranking also depended on selection: for the configuration that pooled validation selects for each model, the differences between models were not significant (Friedman p = 0.57), whereas for the best unseen-house configuration of each model, LightGBM (mean rank 1.6) and XGBoost (2.6) ranked first and the differences were significant ($\chi^{2}$ = 16.29, p = 0.012). This second comparison selects on the test houses and is optimistic, but it suggests that boosted ensembles remain the strongest family once their imbalance strategy is chosen for new homes.

{\footnotesize\setlength{\tabcolsep}{3pt}
\begin{longtable}{>{\raggedright\arraybackslash}p{0.1575\dimexpr\linewidth-36pt\relax}>{\raggedright\arraybackslash}p{0.3384\dimexpr\linewidth-36pt\relax}>{\raggedright\arraybackslash}p{0.0683\dimexpr\linewidth-36pt\relax}>{\raggedright\arraybackslash}p{0.2854\dimexpr\linewidth-36pt\relax}>{\raggedright\arraybackslash}p{0.0683\dimexpr\linewidth-36pt\relax}>{\raggedright\arraybackslash}p{0.0820\dimexpr\linewidth-36pt\relax}}
\caption{Configuration selected by pooled cross-validation and best configuration on unseen houses for each appliance, among all 270 configurations. F1 values are leave-house-out scores; the last column is the F1 lost on unseen houses by selecting with pooled cross-validation. The best configuration on unseen houses is chosen on the test houses and is therefore an optimistic upper bound; Section 4.9 reports a nested house-grouped selection.}\label{tab:10}\\
\toprule
\textbf{Appliance} & \textbf{Selected by pooled CV} & \textbf{F1, unseen houses} & \textbf{Best on unseen houses} & \textbf{F1, unseen houses} & \textbf{F1 lost}\tabularnewline
\midrule
Fridge & RF, FS-B, CW & 0.474 & SVM-RBF, FS-A, ADASYN & 0.477 & +0.003\tabularnewline
Washing machine & XGBoost, FS-B, no correction & 0.474 & LightGBM, FS-B, SMOTE & 0.500 & +0.026\tabularnewline
Dishwasher & XGBoost, FS-B, no correction & 0.224 & XGBoost, FS-B, SMOTE & 0.374 & +0.150\tabularnewline
Kettle & XGBoost, FS-B, no correction & 0.421 & LightGBM, FS-B, SMOTE & 0.479 & +0.058\tabularnewline
Microwave & XGBoost, FS-B, CW & 0.224 & LightGBM, FS-B, SMOTE & 0.243 & +0.019\tabularnewline
\bottomrule
\end{longtable}
}

\subsection{Robustness of the findings}\label{robustness-of-the-findings}

The findings were stable across house partitions. For the four fastest models, repeating the factorial on three further random partitions gave a feature-engineering gain of 55.3\% $\pm$ 1.4\% (mean $\pm$ standard deviation over the four partitions) and a class-weighting gain over no correction of 64.5\% $\pm$ 2.1\%. The share of variance was 4.8\% $\pm$ 0.3\% for the model against 10.2\% $\pm$ 0.2\% for the imbalance strategy, and the rank correlation between pooled and unseen-house scores was 0.91. In every partition, pooled validation selected a configuration that was not the best on unseen houses for four or five of the five appliances (mean 4.25), with a largest loss of 0.143 $\pm$ 0.009, and the support-weighted F1 score of the pooled selections was 0.449 $\pm$ 0.007.

House-grouped tuning did not change the conclusions. Re-tuning LR, DT, XGBoost and LightGBM on the training houses only changed their unseen-house F1 scores by +0.003 on average (mean absolute change 0.008), and the tuned and fixed scores of the 80 configurations were almost perfectly rank-correlated ($\rho$ = 0.99). The benefit of class weighting over no correction was 26\% with tuning and 28\% with fixed hyperparameters on these engineered-feature configurations, and the best tuned configuration used an imbalance correction for four of the five appliances. The optimism introduced by tuning on pooled data is therefore negligible compared with the effects reported above.

Removing the solar-affected Houses 11 and 21 and the House 4 kettle channel did not change the conclusions either (Supplementary Table S15). For the four fastest models, the feature-engineering gain was 55.7\% (56.6\% with all houses) and the class-weighting gain over no correction 57.4\% (61.9\%). The unseen-house scores of the 161 re-run configurations were almost perfectly rank-correlated with those obtained with all houses ($\rho$ = 0.97) and rose by 0.013 on average, and the support-weighted F1 score of the pooled selections rose from 0.447 to 0.459. The best of the four models on unseen houses still used an imbalance correction for every appliance and differed from the pooled selection for all five.

Without calibration, the same-protocol CNN reached a support-weighted F1 score of 0.450 on unseen houses, against 0.447 for the classical configurations selected by pooled validation and 0.467 for the best classical configurations on unseen houses (Table 11). The weighted value is dominated by the fridge; the unweighted means were 0.307 for the CNN, 0.363 for the pooled selections and 0.415 for the best classical configurations. The CNN was competitive for the washing machine (0.490), the fridge (0.468) and the dishwasher (0.344), but weak for the kettle (0.150) and the microwave (0.080). For these two loads it detected more than 92\% of the ON samples at a precision below 0.09, because the class-weighted loss and the fixed threshold of 0.5 placed its operating point far from the F1 optimum; the weakness is one of calibration rather than evidence that the engineered features carry more information. The calibrated variant, whose number of epochs and decision threshold were chosen on validation houses held out from training, confirmed this (Table 11). With the causal window, which like the classical features uses only past samples, it reached a support-weighted F1 score of 0.496 and an unweighted mean of 0.460, above the classical pooled selections (0.447 and 0.363) and the best classical configurations chosen on the test houses (0.467 and 0.415). The centred window gave a similar weighted score (0.497) but a lower mean (0.429). The calibrated causal CNN was better than the classical pooled selections for the washing machine (0.632), the kettle (0.533), the dishwasher (0.390), the fridge (0.497) and, by a small margin, the microwave (0.249). Part of this advantage may come from calibration itself, because the classical models were evaluated at their default threshold. In either form, a deep model evaluated on unseen houses with the present labels and metric does not reproduce the weighted F1 scores of 0.78 to 0.91 listed in the comparison of Pirnat et al. \cite{r12}.

{\footnotesize\setlength{\tabcolsep}{3pt}
\begin{longtable}{>{\raggedright\arraybackslash}p{0.3188\dimexpr\linewidth-36pt\relax}>{\raggedright\arraybackslash}p{0.1362\dimexpr\linewidth-36pt\relax}>{\raggedright\arraybackslash}p{0.1362\dimexpr\linewidth-36pt\relax}>{\raggedright\arraybackslash}p{0.1362\dimexpr\linewidth-36pt\relax}>{\raggedright\arraybackslash}p{0.1362\dimexpr\linewidth-36pt\relax}>{\raggedright\arraybackslash}p{0.1362\dimexpr\linewidth-36pt\relax}}
\caption{Leave-house-out F1 score of the same-protocol CNN, without calibration (fixed threshold of 0.5) and with the number of epochs and the decision threshold chosen on validation houses (centred and causal windows), and of the classical configurations, by appliance. The weighted value is the support-weighted F1 score of Eq. (2); the best classical configuration on unseen houses is chosen on the test houses and is therefore optimistic.}\label{tab:11}\\
\toprule
\textbf{Appliance} & \textbf{CNN, uncalibrated} & \textbf{CNN, calibrated, centred} & \textbf{CNN, calibrated, causal} & \textbf{Classical, selected by pooled CV} & \textbf{Classical, best on unseen houses}\tabularnewline
\midrule
Fridge & 0.468 & 0.492 & 0.497 & 0.474 & 0.477\tabularnewline
Washing machine & 0.490 & 0.663 & 0.632 & 0.474 & 0.500\tabularnewline
Dishwasher & 0.344 & 0.441 & 0.390 & 0.224 & 0.374\tabularnewline
Kettle & 0.150 & 0.390 & 0.533 & 0.421 & 0.479\tabularnewline
Microwave & 0.080 & 0.160 & 0.249 & 0.224 & 0.243\tabularnewline
Weighted F1 & 0.450 & 0.497 & 0.496 & 0.447 & 0.467\tabularnewline
\bottomrule
\end{longtable}
}

\section{Discussion}\label{discussion}

\subsection{Which conclusions survive validation on unseen houses}\label{which-conclusions-survive-validation-on-unseen-houses}

The main conclusions that a pooled benchmark of this design would report fared differently on unseen houses (RQ2). Two conclusions were robust: the appliance dominated the variance under both protocols, and engineered features raised performance by the same 52\% under both. The others changed. The differences between the models selected by pooled validation were no longer significant, the benefit of class weighting more than doubled, an imbalance correction was preferred for most appliances once configurations were selected for unseen houses, and the kettle lost its place as the easiest appliance. These changes are systematic rather than random, and for the four fastest models they were reproduced on four house partitions and under house-grouped tuning: pooled validation preserved the broad ordering of configurations, but, as expected when selection and evaluation share information \cite{r49}, it rewarded configurations that exploit house-specific patterns and penalised imbalance correction, which is precisely the property needed in a new home. For machine-learning methodology, the central implication is that pooled cross-validation, the default protocol of most benchmarks of this kind, did not select the best configuration for unseen homes for any of the five appliances, at a cost of up to 0.150 in F1 (Table 10).

Similar patterns have been reported in other areas of machine learning, where differences between methods shrank under a consistent and stricter protocol \cite{r38,r54}. For NILM, the validation protocol is itself a design factor: selecting configurations with pooled instead of house-grouped validation cost up to 0.110 in F1 on unseen houses in a nested comparison (dishwasher), although the gain over all appliances was small (weighted F1 0.455 against 0.446). Conclusions about which model, features or imbalance strategy to use in new buildings should therefore be drawn from house-grouped validation.

\subsection{Comparison with published results}\label{comparison-with-published-results}

Table 12 places the present results alongside the studies reviewed in Section 2. Among the classical studies, the best F1 scores obtained here (0.324 to 0.570) are comparable to the F1 score of 0.528 attributed to Tabatabaei et al. on REDD \cite{r4,r12} and lower than the micro-averaged F1 scores of 0.68 to 0.70 of Singh and Majumdar \cite{r6} and the F-score of 0.98 of Wu et al. \cite{r7}. The last result was obtained on BLUED, with high-frequency electrical features that cannot be computed at the 8 s resolution of REFIT and with training and test data from consecutive days of a single house, so the difference is likely to reflect the data and the protocol rather than the model. The accuracies of up to 98\% reported by Rehmani et al. \cite{r5} cannot be compared with F1 scores: in the present study, configurations with an accuracy of 0.996 had an F1 score of zero (Section 4.1).

{\footnotesize\setlength{\tabcolsep}{3pt}
\begin{longtable}{>{\raggedright\arraybackslash}p{0.2292\dimexpr\linewidth-30pt\relax}>{\raggedright\arraybackslash}p{0.3197\dimexpr\linewidth-30pt\relax}>{\raggedright\arraybackslash}p{0.1930\dimexpr\linewidth-30pt\relax}>{\raggedright\arraybackslash}p{0.1617\dimexpr\linewidth-30pt\relax}>{\raggedright\arraybackslash}p{0.0964\dimexpr\linewidth-30pt\relax}}
\caption{Comparison with published results. The deep-learning scores of 0.78 to 0.91 are those listed in the comparison of Pirnat et al. \cite{r12}, whose own experiments used an 80:20 split of synthesised data; the value for Tabatabaei et al. is taken from \cite{r12}; the other values are as reported by each study. For Langevin et al. and Tanoni et al., the first value is from the original study and the second is the value listed by Pirnat et al. \cite{r12}, who re-ran the model of Tanoni et al.}\label{tab:12}\\
\toprule
\textbf{Study} & \textbf{Approach} & \textbf{Dataset} & \textbf{Metric} & \textbf{Score}\tabularnewline
\midrule
Tabatabaei et al. \cite{r4} & Multi-label classifiers & REDD & F1 & 0.528\tabularnewline
Singh and Majumdar \cite{r6} & Sparse representation & REDD, Pecan Street & Micro F1 & 0.68 to 0.70\tabularnewline
Wu et al. \cite{r7} & Random Forest, high-frequency features & BLUED (one house) & F-score & 0.98\tabularnewline
Rehmani et al. \cite{r5} & k-NN, RF, Extra-Trees & REDD (one house) & Accuracy & up to 0.98\tabularnewline
Langevin et al. \cite{r11} & Autoencoder-based disaggregation & REFIT & F1; weighted F1 \cite{r12} & 0.76; 0.78\tabularnewline
Tanoni et al. \cite{r10} & Convolutional recurrent network & REFIT, held-out houses; 80:20 split & Micro F1; weighted F1 \cite{r12} & 0.63; 0.83\tabularnewline
Pirnat et al. \cite{r12} & Convolutional transpose recurrent network & REFIT, 80:20 split & Weighted F1 & 0.91\tabularnewline
This study, per appliance & Tuned classical models & REFIT & F1 & 0.324 to 0.570\tabularnewline
This study, five appliances & Tuned classical models & REFIT & Weighted F1 & 0.548\tabularnewline
This study, unseen houses & Best configurations, leave-house-out & REFIT & Weighted F1 & 0.447\tabularnewline
This study, same-protocol CNN & Convolutional network, leave-house-out, calibrated (uncalibrated) & REFIT & Weighted F1 & 0.496 (0.450)\tabularnewline
\bottomrule
\end{longtable}
}

In the comparison of Pirnat et al. \cite{r12}, deep models on REFIT reach weighted F1 scores of 0.78 to 0.91, and the support-weighted F1 score of the present classical models, 0.548 under pooled validation and 0.447 on unseen houses, is between 0.23 and 0.46 lower. The protocol of that study differs from the present one in several respects: the inputs were synthesised from the measurement datasets, the task was multi-label classification of windows of about 6 h, and the 80:20 split is not described as random or grouped by house; appliance subsets, label definitions and training procedures also differ between NILM studies \cite{r40,r41}. The same-protocol CNN (Section 4.10) removes these differences for one deep model: evaluated on the same unseen houses with the same labels and metric, it reached 0.450 uncalibrated and 0.496 calibrated, against 0.447 for the classical models; even the calibrated value is 0.28 to 0.41 below the published deep-learning scores. Under a common protocol, the advantage of a deep model was therefore much smaller than the published gap. This study cannot separate how much of the remaining difference reflects the published protocol and how much the stronger published architectures, which is why NILM methods should be compared only under a common, house-grouped protocol.

Pirnat et al. identified class-imbalance handling as an open question for deep models \cite{r12}. For classical models, imbalance handling improved average performance but could not close a gap of this size: class weighting, the best strategy on average, improved the fridge by 0.220 but reduced performance for three other appliances under pooled validation (Table 6). Although its benefit was larger on unseen houses (Section 4.9), the best unseen-house scores remained far below the published deep-learning results. Newer architectures such as NILMFormer \cite{r29} may widen the gap again, so the comparison should be repeated with stronger deep models.

\subsection{Sources of heterogeneity in classical NILM results}\label{sources-of-heterogeneity-in-classical-nilm-results}

The meta-analysis of Jaime et al. \cite{r15} attributed the spread of classical NILM results to heterogeneity between studies. Within the present single dataset and protocol, the difference between the best and worst of the 54 configurations of each appliance, including configurations that detected almost no ON events, ranged from 0.324 (microwave) to 0.570 (kettle), comparable to the range of 0.45 between the classical studies in Table 12 that report F1 scores. Design choices alone can therefore reproduce the between-study spread, and the variance decomposition (Section 4.7) and the leave-house-out results (Sections 4.8 to 4.10) identify its main sources: the appliance, the feature representation and the validation protocol, which also governs the apparent value of imbalance handling. Two further sources are methodological. Accuracy exceeded F1 in every configuration, so studies that report accuracy appear stronger on the same data. Training-set size changed the best configuration for four of five appliances (Section 4.6), although training-set sizes are often not reported \cite{r7}; both are instances of the undisclosed training conditions that hinder reproducibility \cite{r37} and of benchmarking protocols that create spurious differences between methods \cite{r38}.

The training-set-size effect on RF (Section 4.6) combines two changes: the maximum-depth grid was restricted to 12, and the training data grew from a fixed subsample to the full pooled data. To separate them, RF was re-run on the same fixed subsamples and seed as the first pass with the depth grid restricted to 12. Of the 40 matched configurations, 23 had already selected a bounded depth and were unaffected. The remaining 17 had selected an unrestricted depth, and restricting them to depth 12 reduced their F1 score by 0.091 on average (median $-$0.104). Over all 40 configurations, the depth restriction alone accounts for a mean reduction of 0.039, most of the 0.053 mean decline recorded for RF; the remaining decline of about 0.014 reflects the change to the full pooled data. RF's apparent sensitivity to training-set size is therefore driven mainly by a grid change forced by memory limits, a model-specific source of heterogeneity that a study reporting only headline numbers would not reveal.

\subsection{A reporting checklist for NILM benchmarks}\label{a-reporting-checklist-for-nilm-benchmarks}

Table 13 condenses the findings into a checklist for benchmarks of appliance classification. Each item is tied to an effect measured in this study, so that the cost of omitting it can be judged. Most items apply equally to deep-learning studies, as the same-protocol comparison shows, whose comparisons with classical baselines are affected by the same choices.

{\footnotesize\setlength{\tabcolsep}{3pt}
\begin{longtable}{>{\raggedright\arraybackslash}p{0.3928\dimexpr\linewidth-12pt\relax}>{\raggedright\arraybackslash}p{0.6072\dimexpr\linewidth-12pt\relax}}
\caption{Reporting checklist for NILM appliance-classification benchmarks, with the evidence from this study.}\label{tab:13}\\
\toprule
\textbf{Item} & \textbf{Evidence from this study}\tabularnewline
\midrule
1. Report the F1 score (or another minority-class metric) of the ON class, not accuracy & Accuracy exceeded F1 in all 270 configurations; models with F1 = 0 reached an accuracy of 0.996 (Section 4.1)\tabularnewline
2. Validate on houses unseen during training and state the split & Weighted F1 0.101 lower on unseen houses, and up to 0.288 lower for one appliance (Section 4.8)\tabularnewline
3. Select models, features and imbalance strategies with house-grouped validation & In a nested comparison, house-grouped selection improved F1 on unseen houses by up to 0.110; the pooled choice was not the best on unseen houses for four or five appliances in each of four partitions (Sections 4.9 and 4.10)\tabularnewline
4. Report results per appliance, with variability over folds or houses & The appliance explained 38\% of the variance; test houses explained six times more variance than the model (Sections 4.7 and 4.8)\tabularnewline
5. Report the imbalance strategy for each appliance & Appliance-by-strategy interaction of 14\% of the variance; the best strategy differed between appliances (Sections 4.4 and 4.7)\tabularnewline
6. State the training-set size of every model and test its effect when subsampling & Removing a training-set cap changed the best configuration for four of five appliances, three of them near-ties (Section 4.6)\tabularnewline
7. Report the hyperparameter search space, including changes forced by computing limits & A forced depth restriction explained most of the apparent training-set effect on RF (Section 5.3)\tabularnewline
8. Compare models with a test over appliances or houses, not single scores & Model differences significant under pooled validation (p = 0.002) but not on unseen houses (p = 0.57)\tabularnewline
9. Compare methods, including deep and classical models, under one protocol & Under the same protocol a CNN reached 0.496 against 0.447 for classical models, far below the 0.78 to 0.91 reported under a different protocol (Sections 4.10 and 5.2)\tabularnewline
\bottomrule
\end{longtable}
}

\subsection{Implications for deployment}\label{implications-for-deployment}

The results rank the three design factors by practical value. Feature engineering had the largest and most robust effect (a relative gain of 52\%). Imbalance correction had a modest average effect under pooled validation (32\% for class weighting) but a larger one on unseen houses, where it also changed the best configuration. Model choice mattered least, with XGBoost the most consistent model but without a significant advantage on unseen houses. This ordering favours low-cost deployment: the FS-B features are rolling statistics over 15 samples that can be updated incrementally on streaming meter data, and inference with a trained tree ensemble requires little computation. Training is the costly step, since RF required a depth limit to fit in 18 GB of memory and SVM-RBF could not be trained on the full data, so models would be trained centrally and deployed for inference on the meter or a local gateway. Because a model trained on other homes loses accuracy in a new home, most strongly for the dishwasher (Section 4.8), a deployment should select its configuration by house-grouped validation and prefer an imbalance correction. It should also either fine-tune on a short labelled period in the target home or restrict itself to the appliances that transfer best, the fridge and the washing machine.

The operating points in Table 7 determine which applications are feasible. For the washing machine, the dishwasher and the kettle, the best classical configurations rarely raise false alarms but miss more than half of the ON events, which may be acceptable for coarse usage feedback but not for applications that must capture individual events, such as demand-response schemes triggered by appliance activation. The fridge shows the opposite trade-off. No method tested here, classical or deep, provides reliable event detection for all five appliances in unseen houses, so such applications require further work, for example fine-tuning in the target home.

\subsection{Limitations}\label{limitations}

Several limitations should be acknowledged. The study uses a single dataset, REFIT, so transfer to other datasets and countries was not measured. The leave-house-out experiments cover all 270 configurations on one partition and the four fastest models on four partitions, but each fold tests only three or four houses, so fold-level scores vary considerably. Most leave-house-out results use hyperparameters fixed from the pooled tuning; house-grouped tuning of four models changed their scores by less than 0.01 on average, but it was not repeated for RF, k-NN and SVM-RBF. The CNN was a small network whose architecture was not tuned, so it does not represent the best deep-learning performance attainable under this protocol; conversely, only the CNN had its decision threshold chosen on validation houses, whereas the classical models were evaluated at their default threshold, so part of its advantage may reflect calibration. The data-quality issues of Section 3.2 were retained in the main analysis, and their effect was checked for the four fastest models only (Section 4.10). The nested comparison of selection protocols covers four models with engineered features rather than the full factorial.

The final run is not fully uniform across models: RF was tuned with a maximum depth of 12, an effect that Section 5.3 quantifies, and SVM-RBF remained limited to 100,000 samples. The rolling window of FS-B was fixed at 15 samples and not tuned per appliance, which may not suit short loads such as the kettle and the microwave. The Nemenyi test with five appliances has limited power, and the analysis of variance treats configurations as independent observations although they share data, so its p-values are descriptive. The gain from house-grouped selection was likewise not significant over the 25 folds (p = 0.09), so it is reported as a description of three appliances rather than as a general effect. The support-weighted F1 score is dominated by the fridge, which holds 79

\section{Conclusions and future work}\label{conclusions-and-future-work}

This study asked what determines the performance of classical machine learning for appliance ON/OFF classification, and which of the answers survive validation on houses unseen during training. A factorial experiment of 270 configurations on REFIT, a variance decomposition, a leave-house-out re-evaluation of every configuration, repeated house partitions, house-grouped tuning and selection, and a same-protocol deep baseline give five answers. The study proposes a factorial design with variance decomposition and a leave-house-out selection protocol, and its results apply to appliance-level monitoring for smart-metering and energy-management systems.

First, the task dominates the method: the appliance explained 38\% of the variance in F1, more than the main effects of the feature set (13\%), the model (8\%) and the imbalance strategy (2\%) combined, although imbalance handling also acted strongly through its interaction with the appliance (14\%). Second, feature engineering is the most reliable modelling lever: engineered rolling-window features raised the mean F1 score by 52\% under both pooled validation and validation on unseen houses. Third, pooled validation is a biased guide to model selection for new homes. It preserved the broad ordering of configurations, but it inflated the scores of the best configurations (support-weighted F1 of 0.548 against 0.447 on unseen houses) and concealed the value of imbalance correction, whose benefit rose from 32\% to 69\%. In a nested comparison, selecting configurations by house-grouped validation on the training houses improved F1 on unseen houses for three of five appliances, by up to 0.110, and was within 0.007 for the other two, so the overall gain was small (weighted F1 of 0.455 against 0.446). For the four fastest models, these effects were reproduced on four house partitions and without the solar-affected houses. Fourth, the headline result also depends on how the models were trained: a larger training set raised the best microwave F1 score from 0.256 to 0.324 with the same configuration, and a forced change of the hyperparameter grid explained most of the apparent effect of training-set size on RF. Fifth, under the same protocol a convolutional network scored 0.450 without calibration and 0.496 with a calibrated decision threshold, against 0.447 for the classical models, and far below the 0.78 to 0.91 published under a different protocol, so the published gap cannot be taken as the advantage of deep models under a common protocol.

Classical models with engineered features are therefore an inexpensive option for appliance classification in new homes, with low inference cost on existing meters and gateways. They are within 0.03 to 0.05 in weighted F1 of a calibrated same-protocol deep model (0.496 against 0.447 to 0.467), although the gap is wider in the unweighted mean over the five appliances (0.460 against 0.363 to 0.415), because the weighted score is dominated by the fridge. Detection of the dishwasher and the microwave in unseen houses, however, remained weak for every method tested. For future benchmarks, classical or deep, the results support selecting and reporting models with house-grouped validation, reporting the minority-class F1 score per appliance, stating training-set sizes and comparing methods under one protocol, as summarised in the checklist of Table 13.

Future work will test transfer to other datasets such as REDD \cite{r13} and UK-DALE \cite{r55}, evaluate stronger deep architectures under the same house-grouped protocol, tune the rolling window for each appliance, and examine fine-tuning on a short, labelled period in the target home.

\section*{Declaration of AI Assistance}

During the preparation of this work, the author(s) used Large Language Models (LLMs) to improve language and readability and to help write code for analyses designed by the author(s). The author(s) reviewed and edited all content and code and take full responsibility for the content of the publication.

\ifanon\else
\section*{CRediT authorship contribution statement}

Zablon Ombiro: Conceptualization, Methodology, Software, Validation, Formal analysis, Investigation, Data curation, Visualization, Writing, original draft, Writing, review and editing.
\fi

\section*{Funding}

This research did not receive any specific grant from funding agencies in the public, commercial, or not-for-profit sectors.

\ifanon\else
\section*{Declaration of competing interest}

The author declares that he has no known competing financial interests or personal relationships that could have appeared to influence the work reported in this paper.
\fi

\section*{Data and code availability}

The REFIT dataset is public and openly available from the University of Strathclyde repository \cite{r21}; the raw data are not redistributed with this paper or with the code and must be obtained from the original source. The source code, the verified appliance-channel map, the verification script, the scripts that regenerate the feature tables from the raw data and the complete results of all experimental runs are available at \ifanon\else\url{https://github.com/IncognitoGK9/non\_intrusive\_load\_management}\fi, so that every result can be replicated from public data. The per-house data audit and the results tables are also provided as supplementary material.

\section*{Supplementary material}

The Supplementary Material contains Supplementary Tables S1 to S16: the per-house data audit (S1), the first-pass and final results (S2, S3), the matched comparison of the 160 re-run configurations (S4), the Random Forest depth-control experiment (S5), the leave-house-out validation and factorials (S6 to S9, including the analysis-of-variance tables), the repeated house partitions (S10), the house-grouped tuning (S11), the same-protocol and calibrated CNN (S12, S14), the nested comparison of pooled and house-grouped selection (S13), the data-quality sensitivity analysis (S15) and the computing environment and run times (S16).

\end{document}